\pdfoutput=1

\documentclass[11pt]{article}
\usepackage[final]{acl}

\usepackage{booktabs}
\usepackage{multirow}
\usepackage{makecell}
\usepackage{graphicx}
\usepackage{xspace}
\usepackage{enumitem}
\usepackage[most]{tcolorbox}
\usepackage{inconsolata}

\tcbuselibrary{skins}

\definecolor{promptVerify}{HTML}{2F5D8C}   
\definecolor{promptExtract}{HTML}{1F7A6B}  
\definecolor{promptLabel}{HTML}{8A4B7C}    
\definecolor{promptTint}{HTML}{FAFBFD}

\tcbset{
  promptbox/.style 2 args={
    enhanced,
    colback=promptTint,
    colframe=#1,
    coltitle=white,
    colbacktitle=#1,
    title=#2,
    fonttitle=\bfseries\small,
    fontupper=\small,
    boxrule=0.8pt, titlerule=0pt, arc=3pt,
    left=7pt, right=7pt, top=5pt, bottom=5pt,
    toptitle=2pt, bottomtitle=2pt,
    width=\linewidth,
  }
}
\newcommand{\system}{\textsc{Peerify}\xspace}

\title{\system: Benchmarking Peer-Review Claim Verification}

\author{
 \textbf{Alireza Daghighfarsoodeh\textsuperscript{1}},
 \textbf{Sajad Ebrahimi\textsuperscript{1,2}},
 \textbf{Ali Ghorbanpour\textsuperscript{1}},\\
 \textbf{Soroush Sadeghian\textsuperscript{1}},
 \textbf{Radin Cheraghi\textsuperscript{1}},
 \textbf{Negar Arabzadeh\textsuperscript{1,3}},\\
 \textbf{Ebrahim Bagheri\textsuperscript{1,2}}
\\
\\
 \textsuperscript{1}Reviewerly,
 \textsuperscript{2}University of Toronto,
 \textsuperscript{3}University of California, Berkeley
\\
{ 
   \textbf{Correspondence:} \href{mailto:s.ebrahimi@utoronto.ca}{s.ebrahimi@utoronto.ca}
   }
\\
 \texttt{\{daqiq, aligh, soroush, radin, arabzadeh\}@reviewerly.ca},\\
 \texttt{\{s.ebrahimi, ebrahim.bagheri\}@utoronto.ca}\\
}

\begin{document}
\maketitle

\begin{abstract}
Peer review plays a central role in scholarly publishing, yet verifying whether reviewer claims are supported by manuscript evidence remains a largely manual and time-consuming process. We present \system, a pipeline for manuscript-grounded verification of peer-review claims. Given a manuscript and a review comment, \system pipeline decomposes reviews into atomic claims, retrieves relevant manuscript evidence, and determines whether each claim is supported by the paper. To support the development and evaluation of the pipeline, we construct a benchmark of 800 claims derived from authentic peer-review interactions collected from NeurIPS 2024 and ICLR 2024, including a 300-claim hand-labeled subset used to audit the automated supervision. We evaluate state-of-the-art language models and retrieval strategies within \system pipeline, together with entailment baselines. Our results demonstrate the importance of retrieval-centered verification and claim decomposition, while highlighting the challenges posed by ambiguous and interpretive reviewer claims. Automated labels agree with human consensus on 90.3\% of audited claims ($\kappa = 0.87$), while off-the-shelf entailment models stay below 0.24 macro-F1. 

\end{abstract}

\section{Introduction}

Peer review plays a central role in how scientific knowledge is evaluated and incorporated into the scholarly record 
and help establish confidence in scientific findings \citep{lee2013peer,bornmann2011peer,smith2006peer,arabzadeh2024reviewerly}. 
As submission volumes grow, editors, area chairs, and program committees must assess more reviews under tighter timelines.
Maintaining review quality and consistency at scale has therefore become an important operational challenge for scholarly publishing \citep{ebrahimi2025exharmony, Mulligan2013PeerRI, arabzadeh2025building, ebrahimi2026peeriscope}.

A substantial portion of the review process relies on claims made by reviewers which often influence publication outcomes. 
However, verifying whether such statements are supported by the contents of the manuscript remains largely a manual process. In practice, editorial stakeholders rarely have time to systematically examine every review statement against the corresponding paper. 
Consequently, unsupported claims and misunderstandings of a manuscript may persist throughout the review process without systematic verification \cite{ghorbanpour2026peerispect}.

The growing adoption of Large Language Models (LLMs) within scholarly workflows further amplifies the importance of this problem \citep{bender2021dangers, arabzadeh2026can}. LLMs are increasingly used to assist with manuscript preparation, review drafting, and editorial support tasks \citep{liang2024feedback,yuan2022reviewgen,sadeghian2026peerprism}. While these technologies offer substantial opportunities for improving efficiency, they also increase the need for mechanisms that can assess whether statements contained in reviews are faithfully grounded in the manuscript \citep{maynez2020faithfulness,huang2025survey}. Reliable verification of review claims is therefore becoming an increasingly important component of research integrity 
within modern publishing ecosystems.

Although claim verification has been extensively studied in NLP \citep{dagan2006rte,thorne2018fever,wadden2020scifact}, peer-review claim verification differs in both scope and motivation. Existing work on LLM-assisted peer review focuses on review generation or quality assessment \citep{yuan2022reviewgen,liang2024feedback,hua2019argument,rogers2020peerreview,ebrahimi2025rottenreviews}. However, to the best of our knowledge, verifying whether individual reviewer claims are grounded in the manuscript remains underexplored.
Scientific claims frequently depend on methodological assumptions, dataset characteristics, and experimental conditions that must be interpreted together rather than in isolation. Furthermore, review statements often combine objectively verifiable assertions with subjective judgments, requiring systems to distinguish between components that can be grounded in evidence and those that reflect reviewer opinion \citep{metropolitansky2025claimify,pavlick2019disagreements,plank2022problem}. Operationalizing review groundedness thus requires robust document understanding, evidence retrieval, multi-hop reasoning, and careful treatment of ambiguity  \citep{pavlick2019disagreements,plank2022problem,arabzadeh2026doxa}. We provide a more comprehensive discussion of related work in Appendix~\ref{app:related}.

We identify manuscript-grounded peer-review claim verification as an important but underexplored challenge in scholarly publishing. 
To address this challenge, we propose \system, an end-to-end pipeline for peer-review claim verification. \system pipeline decomposes reviews into atomic, self-contained claims, retrieves relevant manuscript evidence for each claim, and performs groundedness assessment. \system pipeline was designed with practical deployment requirements in mind, including scalability to large review volumes, transparent evidence attribution, robustness across diverse claim types, and compatibility with existing editorial processes.
We use \emph{production-ready} in an engineering sense: the pipeline is complete, runs on commodity hardware, and is designed for a human-in-the-loop workflow in which \system\ triages reviewer claims and surfaces the supporting passages, and an area chair or author acts on them (Appendix~\ref{app:deployment}).

To systematically evaluate this task and pipeline, we construct a benchmark of 800 atomic review claims derived from publicly available NeurIPS~2024 and ICLR~2024 submissions, together with their reviews and author--reviewer discussion threads.  The benchmark captures diverse verification scenarios, including explicitly supported claims, claims requiring evidence aggregation, ambiguous claims, and claims that cannot be resolved from the manuscript alone.

Using this benchmark, we evaluate \system pipeline end to end and analyze each of its core components, including claim extraction, evidence retrieval, retrieval configurations, and state-of-the-art language-model verifiers. Our results show that retrieval-centered verification and claim decomposition are important for manuscript-grounded review verification, but also reveal that ambiguous, interpretive, and partially supported reviewer claims remain challenging for current methods. We release \system pipeline, the benchmark, labels, prompts, and code to support reproducible research at \url{https://github.com/Reviewerly-Inc/Peerify/}.



\system makes the following contributions:

\begin{enumerate}[leftmargin=*, itemsep=0pt, topsep=0pt, parsep=0pt]

\item \textbf{Task.} We formalize manuscript-grounded peer-review claim verification as an intra-document verification task, where claims from peer reviews are assessed strictly against the reviewed manuscript.

\item \textbf{System.} We propose \system, an end-to-end verification pipeline that decomposes reviews into atomic claims, retrieves relevant manuscript evidence, and predicts explicit evidence attribution.

\item \textbf{Benchmark and evaluation.} We construct an 800-claim benchmark from NeurIPS~2024 and ICLR~2024 peer-review interactions, with six evaluation variants and quality-controlled slices to evaluate \system pipeline, including a 300-claim human-audited subset that covers 37.5\% of the benchmark.


\end{enumerate}

\section{\system Pipeline}
\label{sec:peercheck}
\label{sec:dataset}

\subsection{Problem Definition}
\label{sec:task-formulation}

Let $P$ denote a scientific manuscript and $R$ denote an associated peer review. The objective of review-groundedness verification is to determine whether verifiable claims contained within $R$ are supported by evidence present in $P$. Formally, a review $R$ can be represented as a collection of atomic claims $C=\{c_1,\ldots,c_m\}$. For each claim $c_i$, the system identifies a set of supporting evidence candidates $E_i=\{e_1,\ldots,e_k\}$ from the manuscript and predicts a groundedness label $y_i=f(c_i,E_i)$.

This formulation differs from traditional fact verification and textual entailment settings \citep{dagan2006rte,thorne2018fever,wadden2020scifact} in several important respects. First, the evidence space is restricted to a single scientific manuscript rather than an external corpus. 
Second, evidence supporting a review claim may be distributed across multiple sections, tables, appendices, and more. 
Third, review claims frequently involve methodological choices, experimental design decisions, and interpretations of empirical findings whose validity depends on contextual qualifiers and evidence aggregation. 




\subsection{Operationalization in \system}

\system\ operationalizes review-groundedness verification as a three-stage pipeline:

\noindent \textbf{Step 1: Claim extraction.}
Review comments often contain several assertions in a single sentence, mixing factual observations with subjective judgments or recommendations. Direct verification is difficult because different portions may require distinct evidence. \system\ therefore adopts a claim decomposition strategy, breaking each review comment into atomic, self-contained claims that can be verified independently. This step preserves the reviewer’s intended meaning while isolating factual assertions to be checked against the manuscript.


\noindent \textbf{Step 2: Evidence retrieval.}
The manuscript is segmented into retrieval units such as paragraphs, figure captions, tables, and appendix passages. Given a claim, \system\ retrieves the top-$k$ relevant passages based on semantic similarity between the claim and indexed manuscript segments. Since supporting evidence may be distributed across multiple locations, \system\ retrieves a set of evidence candidates rather than a single passage, allowing downstream verification to reason over complementary evidence throughout the document.

\noindent \textbf{Step 3: Groundedness assessment.}
Finally, \system\ verifies each claim against the retrieved evidence, building on evidence-based verification paradigms from textual entailment, fact verification, and scientific claim verification \citep{dagan2006rte,thorne2018fever,wadden2020scifact}. The verifier jointly reasons over the claim and manuscript evidence to determine the extent of support, assigning one of four labels: \textit{Supported}, \textit{Partially Supported}, \textit{Not Supported}, or \textit{Not Determined}. Each prediction is paired with the evidence used for the decision, making the output transparent and traceable for human inspection.



\begin{table*}[ht]
\centering

\caption{\system\ benchmark statistics overview.}
\vspace{-0.5em}
\resizebox{0.95\textwidth}{!}{%
\begin{tabular}{lcc | cccc | cccc}
\toprule
\textbf{Dataset} & \textbf{\#Papers} & \textbf{\#Claims} &  & \textbf{Partially} & \textbf{Not} & \textbf{Not} & \textbf{ICLR} & \textbf{NeurIPS} & \textbf{Reject} & \textbf{Accept} \\

 &  &  &  \textbf{Supported} & \textbf{Supported} & \textbf{Supported} & \textbf{Determined} &  &  &  &  \\
\midrule
NeurIPS 2024 & 4236 & - & - & - & - & - & - & 100\% & 86.0\% & 14.0\% \\
ICLR 2024 & 5778 & - & - & - & - & - & 100\% & - & 31.3\% & 68.7\% \\
\midrule
\textsc{PaperSourced} (\ref{sec:papersourced}) & 100 & 500 & 500 & 0 & 0 & 0 & 52.0\% & 48.0\% & 27.0\% & 73.0\% \\
\textsc{RebuttalSourced} (\ref{sec:rebuttal}) & 84 & 800 & 217 & 310 & 189 & 84 & 61.0\% & 39.0\% & 34.9\% & 65.1\% \\
\textsc{LLMJudged} (\ref{sec:llmjudge}) & 84 & 800 & 207 & 93 & 141 & 359 & 61.0\% & 39.0\% & 34.9\% & 65.1\% \\
\textsc{HighAgreement} (\ref{sec:qualityslice}) & 71 & 175 & 60 & 39 & 43 & 33 & 63.4\% & 36.6\% & 37.1\% & 62.9\% \\
\textsc{verifiable-only} (\ref{sec:qualityslice}) & 67 & 131 & 56 & 29 & 25 & 21 & 64.9\% & 35.1\% & 36.6\% & 63.4\% \\
\textsc{Human-Verified} (\ref{sec:humanjudge}) & 75 & 300 & 77& 107 & 78 & 38 & 57.3\% & 42.7\% & 34\% & 66\%\\
\bottomrule
\end{tabular}
}
\label{tab:peercheck-distributions}
\end{table*}

\section{Benchmarking \system}

The \system\ pipeline requires an evaluation framework that reflects the challenges of real peer-review workflows. Existing claim-verification resources such as FEVER \citep{thorne2018fever} and SciFact \citep{wadden2020scifact} evaluate claims against external corpora, but do not capture manuscript-bounded verification, author--reviewer interactions, or the mix of factual and subjective assertions common in peer review (see Appendix~\ref{app:related}). We therefore develop the \system\ benchmark, which operationalizes review groundedness as an intra-document verification task and supports the development, validation, and evaluation of review-groundedness verification systems.
To ensure ecological validity, all claims are derived from authentic peer-review interactions and evaluated against the manuscripts under review. The resulting benchmark contains 800 review-derived claims from publicly available NeurIPS~2024 and ICLR~2024 submissions. This is at or above standard expert-annotated claim-verification sets, with every label grounded in a manuscript (Appendix~\ref{app:size}).

\subsection{Source Collection}

We collected manuscripts, reviews, author responses, and discussion threads from publicly available submissions in NeurIPS 2024 and ICLR 2024. These venues provide complete review artifacts, including reviewer comments and author rebuttals, so each instance pairs a manuscript with its review and discussion history and supports multiple sources of evidence about a claim's validity. 
{As reported in Table~\ref{tab:peercheck-distributions}, the collected papers are well balanced across review outcomes, covering both accepted and rejected submissions.}
\subsection{Claim Construction}

Review comments contain summaries, questions, recommendations, methodological critiques, and subjective opinions. Since review-groundedness verification operates at the level of individual factual assertions, raw review text was transformed into atomic claims suitable for independent verification.

To construct these claims, we evaluated three extraction approaches: \texttt{Qwen3-4B} \citep{yang2025qwen3}, prompted to produce atomic, decontextualized assertions; Fenice \citep{scire2024fenice}, which uses dependency parsing and semantic role labeling to identify proposition structures; and Gemma, a general-purpose instruction-tuned model prompted to list atomic claims from each passage.

{We run a controlled comparison of the three extractors and adopt \texttt{Qwen3-4B} for all benchmark construction, as it gives the best balance of atomicity, decontextualization, and semantic fidelity. The full evaluation protocol and results are deferred to Appendix~\ref{app:extraction}.}
We also scored \texttt{o4-mini} and \texttt{GPT-5-mini} as extractors: both are stronger on F1 (0.895 and 0.915 against 0.707), but the gap is mostly recall rather than precision (0.889), and extraction is the high-volume stage where API cost and rate limits bite (Appendix~\ref{app:extract-cost}). A human audit of the semantic-equivalence judge behind these scores gives 89.3\% agreement (Appendix~\ref{app:judge-human}).
Extracted claims were subsequently post-processed through exact and near-duplicate removal, string normalization, and filtering of rhetorical questions, speculative statements, and non-verifiable meta-commentary. 

\subsection{Groundedness Annotation}

{Each extracted claim is assigned one of the four labels of \textit{Supported}, \textit{Not Supported}, \textit{Partially Supported} or  \textit{Not Determined}.}
{Each claim is additionally tagged as \emph{factual} or \emph{subjective} (automatically, with \texttt{o4-mini}; criteria in Appendix~\ref{app:factual}), and the \textsc{verifiable-only} slice retains only factual claims. This ensures groundedness is evaluated on assertions checkable against the manuscript rather than on inherently subjective reviewer opinions.}
The split is close to even (444 objective, 356 subjective), and objective claims counter-intuitively need a human tie-breaker \emph{more} often, 41.4\% against 38.2\% (Appendix~\ref{app:claimtype}).

\subsection{Evaluation Variants}
\label{sec:papersourced}
\label{sec:rebuttal}
\label{sec:llmjudge}
\label{sec:humanjudge}
\label{sec:qualityslice}

\system\ includes four core benchmark variants and two quality-controlled slices, capturing complementary notions of groundedness and label confidence.

\noindent{\textbf{\textsc{PaperSourced}.}}
\textsc{PaperSourced} provides a controlled diagnostic setting in which claims are extracted directly from manuscript content. 
These claims are \emph{verifiable by construction} and are expected to be labeled \texttt{Supported}. We use this subset as a sanity check to isolate errors in retrieval and verification from ambiguity in review-derived content.

\noindent{\textbf{\textsc{RebuttalSourced}.}}
This variant derives supervision from realistic author--reviewer discussion threads, assigning each claim a label by interpreting the author response signal (agreement, correction, clarification, or explicit dispute). {Two language models (\texttt{GPT-5-mini} and \texttt{Claude-Sonnet-4-6}) label each claim independently from the discussion context, with a human arbitrating disagreements to produce the canonical label (Appendix~\ref{app:disagreement}).} This {interaction-grounded} supervision, captures how authors themselves judge the correctness of reviewer claims.

\noindent{\textbf{\textsc{LLMJudged}.}}
This subset provides scalable supervision by judging each claim directly against the manuscript. The same two labelers (\texttt{GPT-5-mini} and \texttt{Claude-Sonnet-4-6}) independently assign one of four labels under strict document-bounded evidence constraints, and a human annotator arbitrates the cases where they disagree to produce the canonical label (More details in Appendix~\ref{app:disagreement}). We evaluate label quality against the \textsc{Human-Verified} labels in {Appendix~\ref{app:agreement}.}

\noindent{\textbf{\textsc{Human-Verified}}}
To provide high-fidelity reference labels and audit automated supervision, we manually annotated a subset of 300 \textsc{RebuttalSourced} instances, which is 37.5\% of the benchmark.
Two annotators with formal training in computer science independently labeled each instance  following detailed guidelines with concrete examples. Inter-annotator reliability was near-perfect (Cohen's $\kappa \approx 0.96$) for the four-way nominal task, with disagreements resolved through consensus.
Against these labels the automated supervision is correct on 90.3\% of the 300 claims under strict four-way exact match ($\kappa = 0.87$), 97.0\% allowing a one-step difference, with hard polarity flips in 1.0\% of cases. Reliability is flat across claim types, 90.5\% objective against 90.1\% subjective, which is where a labeler-inherited bias would surface first (Appendix~\ref{app:human-agreement}).

\begin{figure}[t]
  \centering
  \vspace{-1em}
  \includegraphics[width=\linewidth]{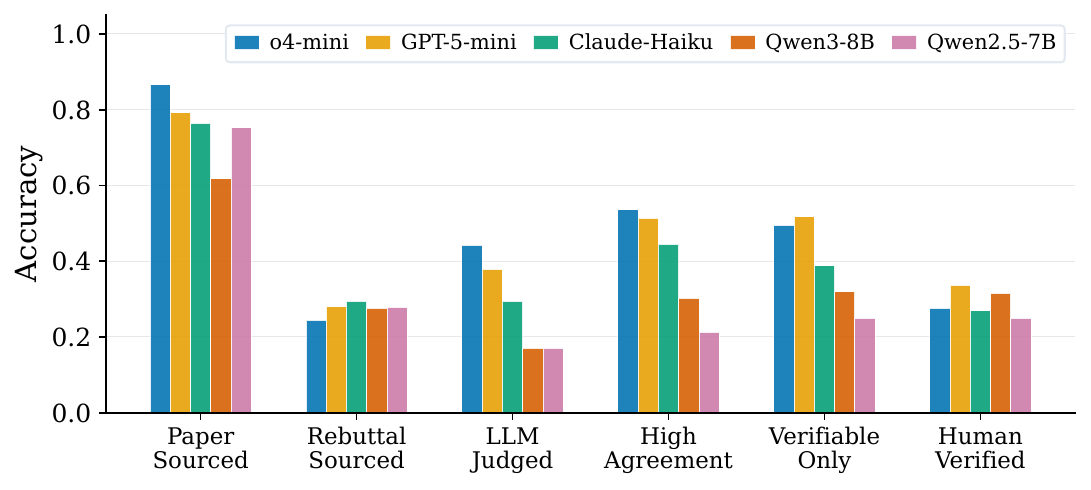}
  \caption{Full-context verification accuracy across  all six \system benchmark variants.}
  \vspace{-1em}
  \label{fig:rq1-accuracy}
\end{figure}

\noindent{\textbf{Quality-Controlled Slices}}
we additionally release slices that improve label reliability. 
First, the \textsc{HighAgreement} slice retains only instances where two supervision signals with disjoint evidence, \textsc{RebuttalSourced} and \textsc{LLMJudged}, agree, yielding a high-confidence subset less sensitive to labeling noise.
Both signals come from the same two labeler models, so a match is not annotator consensus; what differs is the evidence, since \textsc{RebuttalSourced} reads only the discussion thread and \textsc{LLMJudged} only the manuscript, and agreement across disjoint inputs signals claim clarity (see Limitations). Second, to isolate assertions checkable directly against the manuscript, the \textsc{verifiable-only} slice uses \texttt{o4-mini} to filter out subjective or normative claims such as novelty, significance, or impact. Specifically, the model predicts whether a claim is (i) grounded in factual content observable in the manuscript, such as missing experiments or absent figures, or (ii) an opinion, recommendation, or forward-looking judgment not directly verifiable from the paper alone.
Table~\ref{tab:peercheck-distributions} shows the support-level distributions across different variants. \textsc{PaperSourced} is entirely \texttt{Supported} (500/500) by construction and serves as a controlled diagnostic. \textsc{RebuttalSourced} is far more heterogeneous (217 Supported, 310 Partially Supported, 189 Not Supported, 84 Not Determined), reflecting corrections, and unresolved disagreement in author--reviewer exchanges. \textsc{LLMJudged} shifts heavily toward \texttt{Not Determined} (207 Supported, 93 Partially Supported, 141 Not Supported, 359 Not Determined), indicating that many review statements cannot be decisively validated from the manuscript alone.
\section{Experimental setup}

\begin{table*}[t]
\centering

\caption{RAG accuracy across five models and four retrievers on all six \system\ benchmarks. }
\resizebox{0.85\textwidth}{!}{%
\begin{tabular}{llcccccc}
\toprule
\textbf{Model} & \textbf{Retriever}
  & \makecell{\textsc{Paper}\\\textsc{Sourced}}
  & \makecell{\textsc{Rebuttal}\\\textsc{Sourced}}
  & \makecell{\textsc{LLM}\\\textsc{Judged}}
  & \makecell{\textsc{High}\\\textsc{Agreement}}
  & \makecell{\textsc{Verifiable}\\\textsc{-only}}
  & \makecell{\textsc{Human}\\\textsc{Verified}} \\
\midrule
\multirow{4}{*}{\texttt{o4-mini}}
  & BM25    & 0.818          & 0.234          & 0.491          & \textbf{0.514} & 0.466          & 0.243\\
  & {BM25+Reranker}  & 0.870          & 0.234          & 0.496          & \textbf{0.514} & \textbf{0.519} & 0.237          \\
  & Dense   & 0.852          & 0.244& \textbf{0.514} & 0.509          & 0.466          & 0.230          \\
  & {Dense+Reranker} & \textbf{0.904} & 0.240          & 0.499          & 0.497          & 0.466          & 0.257          \\
\midrule
\multirow{4}{*}{\texttt{GPT-5-mini}}
  & BM25    & 0.796          & 0.289          & 0.412          & 0.509& 0.496          & 0.273          \\
  & {BM25+Reranker}  & 0.856& 0.294& 0.417& 0.491          & 0.496          & 0.290          \\
  & Dense   & 0.820          & 0.294& 0.414          & 0.503          & 0.511& 0.297 \\
  & {Dense+Reranker} & 0.852          & 0.286          & 0.405          & 0.491          & 0.466          & 0.257          \\
\midrule
\multirow{4}{*}{\texttt{Claude-Haiku}}
  & BM25    & 0.786          & 0.280          & 0.381          & 0.440          & 0.389          & 0.270          \\
  & {BM25+Reranker}  & 0.822          & 0.270          & 0.372          & 0.451          & 0.405          & 0.240          \\
  & Dense   & 0.808          & 0.256          & 0.386& 0.429          & 0.374          & 0.237          \\
  & {Dense+Reranker} & 0.844& 0.297& 0.374          & 0.480& 0.427& 0.300\\
\midrule
\multirow{4}{*}{\texttt{Qwen3-8B}}
  & BM25    & 0.744          & 0.274          & 0.393& 0.440& 0.382          & 0.257          \\
  & {BM25+Reranker}  & 0.816& 0.281          & 0.364          & 0.440& 0.389& 0.297          \\
  & Dense   & 0.800          & 0.287& 0.371          & 0.400          & 0.321          & 0.283          \\
  & {Dense+Reranker} & 0.814          & 0.265          & 0.385          & 0.423          & 0.374          & 0.277\\
\midrule
\multirow{4}{*}{\texttt{Qwen2.5-7B}}
  & BM25    & 0.844          & 0.319          & 0.211          & 0.263          & 0.267          & 0.300          \\
  & {BM25+Reranker}  & 0.872& 0.323          & 0.211          & 0.297& 0.305& \textbf{0.340} \\
  & Dense   & 0.832          & \textbf{0.324} & 0.209          & 0.263          & 0.260          & 0.307          \\
  & {Dense+Reranker} & 0.860          & 0.295          & 0.228& 0.291          & 0.305& 0.287          \\
\bottomrule
\end{tabular}%
}
\vspace{-1em}
\label{tab:rag-b2-b6}
\end{table*}


\label{sec:baselines}




We evaluate \system\ in two deployment configurations. \emph{Full-Context} feeds the whole manuscript and claim to the verifier, serving as an upper bound when the paper fits within the context window; over-length papers are excluded per model. \emph{Retrieval-Augmented} (RAG) retrieves evidence per claim and conditions the verifier only on the top-$k$ passages, with or without a cross-encoder reranker, enabling deployment on long manuscripts. All prompts are listed in Appendix~\ref{app:prompts-grp}.





{We instantiate \system pipeline with five state-of-the-art models spanning proprietary and open-weight families: \texttt{o4-mini} (200K-context reasoning model), \texttt{GPT-5-mini} (compact, long-context), \texttt{Claude-Haiku-4.5} (low-latency, 200K context), and the open-weight \texttt{Qwen3-8B} and \texttt{Qwen2.5-7B}, the last two evaluated mainly under RAG due to context-window limits. All models receive identical label definitions and prompts, and outputs are mapped to the four support level labels.}






{We evaluate retrieval with recall@$k$, nDCG@$k$, and MRR on \textsc{PaperSourced} (BM25 or a dense retriever, optionally reranked by a cross-encoder; retriever and reranker checkpoints, top-$k$, and chunking are detailed in Appendix~\ref{app:retrieval}), and verification with Accuracy and Macro-F1, taking Macro-F1 as the primary metric since it is robust to label imbalance.} Retrieval metrics are reported only on \textsc{PaperSourced} because gold evidence spans exist only there, so they are inapplicable to the review-derived variants by construction rather than omitted (Appendix~\ref{app:retrieval-scope}).

\noindent\textbf{Non-LLM baselines.} We add three zero-shot MNLI entailment baselines \citep{liu2019roberta,lewis2020bart,he2021deberta}, with the top-3 retrieved chunks as premise and the claim as hypothesis, run under all four retrieval configurations so a weak result cannot be blamed on one retriever (Appendix~\ref{app:nli}).

\section{Results and Findings}
\label{sec:results}
\label{sec:rag_Res}
\label{sec:full_context_resuls}

In this section, we evaluate \system pipeline from several complementary perspectives.
\subsection{Full-context Verification }
Figure~\ref{fig:rq1-accuracy} reports verification accuracy across the six evaluation variants, revealing a clear difficulty gradient. Performance is highest on \textsc{PaperSourced}, where claims come directly from manuscript content and evidence is explicit; \texttt{o4-mini} reaches 0.866, and four of five verifiers exceed 0.75.
Accuracy drops on review-derived settings such as \textsc{LLMJudged} and \textsc{HighAgreement}, where claims require reasoning over dispersed evidence and reviewer intent. Even the strongest verifier reaches only 0.442 on \textsc{LLMJudged}, showing that review-groundedness verification goes beyond simple evidence localization. The hardest setting is \textsc{RebuttalSourced}, where claims come from real author--reviewer interactions and often mix factual observations. \textsc{Human-Verified} behaves the same way, with no verifier clearing 0.38 across its 300 hand-labeled claims.


\subsection{Retrieval-Centered Verification.}
A key design decision in \system\ is separating evidence retrieval from groundedness assessment. Table~\ref{tab:rag-b2-b6} evaluates four retrieval configurations, including BM25, dense retrieval, and their reranked variants. Retrieval quality substantially affects verification accuracy, but no retriever dominates across all settings. On \textsc{PaperSourced}, where evidence is explicit, reranking consistently improves performance. Dense+Reranker increases \texttt{o4-mini} from 0.852 to 0.904, and BM25+Reranker increases \texttt{GPT-5-mini} from 0.796 to 0.856. On review-derived benchmarks, however, gains are less consistent. For example, Dense retrieval performs best for \texttt{o4-mini} on \textsc{LLMJudged} (0.514), while BM25-based retrieval works best for \texttt{GPT-5-mini} (0.417) and \texttt{Qwen3-8B} (0.393).
Similar variability is observed on \textsc{HighAgreement} and \textsc{Human-Verified}, indicating that retrieval effectiveness depends not only on evidence quality but also on how each verification engine consumes retrieved context. Better retrieval therefore helps most when the task requires explicit evidence localization, and much less on \textsc{RebuttalSourced} and \textsc{Human-Verified}, where claims require interpreting reviewer intent and resolving partial support: retrieval is necessary but not sufficient.

\begin{figure}[t]
  \centering
  \includegraphics[width=\linewidth]{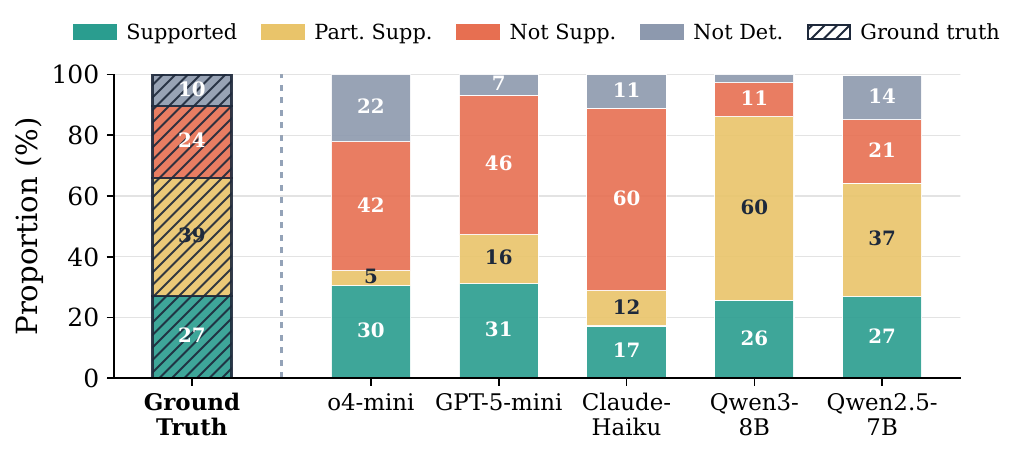}
  \caption{Predicted vs.\ ground-truth (GT) labels on \textsc{RebuttalSourced}. Models exhibit different calibration biases: some over-predict \textit{Not Supported}, while others over-predict \textit{Partially Supported} or better match the GT distribution.}
  
  \label{fig:pred-dist}
\end{figure}
\noindent\textbf{Zero-shot entailment models reach at most 0.24 macro-F1, roughly half the weakest LLM verifier.} Swapping the LLM verifier for an off-the-shelf NLI model, everything else fixed, does not come close: across three MNLI models and all four retrievers, macro-F1 never exceeds 0.24 on any variant, against 0.45 to 0.50 for the best LLM configuration on the document-decidable splits (Appendix~\ref{app:nli}). Two structural causes explain this: three-way NLI cannot express \texttt{Partially Supported}, the largest class in \textsc{RebuttalSourced}, and a 512-token premise limit truncates most of the retrieved evidence. The retriever moves these numbers by at most two points, so the shortfall belongs to the formulation, which needs long-context reasoning.

\subsection{Error Analysis}
\label{sec:error-analysis}

The largest error source is \textsc{RebuttalSourced}, where reviewer claims mix factual observations with interpretive judgments, and calibration differences across verification engines (Figure~\ref{fig:pred-dist}) show that calibration matters as much as reasoning capability. A second failure mode emerges on \textsc{LLMJudged} and \textsc{HighAgreement}, where verification requires multi-step reasoning over dispersed evidence, limiting even the strongest verifier to 0.442 and 0.536 accuracy. The primary bottlenecks are therefore reasoning over distributed evidence and handling interpretive or partially supported claims, a diagnosis confirmed by full-context verification, which removes retrieval entirely and still does not lift accuracy on these splits (Appendix~\ref{app:retrieval-scope}).

\noindent\textbf{Nine in ten \textsc{PaperSourced} errors are strictness or retrieval, not faulty reasoning.} \textsc{PaperSourced} is the one split where the gold label is known for every instance without appeal to a judge, which makes it the right place to ask what a residual error actually consists of. We manually inspected a stratified sample of 150 errors, 30 per verifier under RAG, and assigned each to one of three categories that proved exhaustive. Overly strict specificity accounts for 83 errors (55.3\%): the verifier retrieves the correct passage and then declines to call the claim \texttt{Supported} because one fine detail, an exact count or an appendix pointer, is not restated verbatim. Retrieval misses account for 54 (36.0\%): the supporting passage never enters the top-$k$ chunks, so the verifier correctly reports that the evidence in front of it does not contain the claim. The remaining 13 (8.7\%) are malformed outputs that cannot be parsed into a label.

\section{Conclusion}

We presented \system, a pipeline and benchmark for verifying peer reviews. Our evaluation shows that verifying interpretive claims requires long-context, four-way reasoning rather than simple entailment. Furthermore, retrieval and reasoning create distinct bottlenecks depending on the claim, and raw accuracy is often misleading, making macro-F1 a necessary evaluation metric.
Our automated labeling proved highly reliable, matching human consensus on 90.3\% of an audited subset. Designed for a human-in-the-loop editorial workflow, \system\ accelerates review verification by triaging questionable claims and surfacing manuscript evidence. Future extensions include verifying claims against external literature, joint training, and exploring abstention-aware objectives. We release the pipeline, benchmark, and all artifacts to support this work.

\newpage
\section*{Limitations}

\system\ performs \textbf{intra-document} verification, assessing review claims only against the corresponding manuscript; claims requiring external literature or background knowledge are out of scope. The benchmark is built from public OpenReview data for NeurIPS~2024 and ICLR~2024, so its distribution may not transfer to other fields or venues with different review norms. Some variants rely on weak supervision (author--reviewer discussions and LLM judgments); we mitigate this with the \textsc{HighAgreement} and \textsc{verifiable-only} slices, but residual ambiguity in review language remains. Finally, models show strong calibration bias (e.g., \texttt{Qwen3-8B} predicts \texttt{Partially Supported} for 60.5\% of \textsc{RebuttalSourced} claims vs.\ a 38.8\% ground-truth rate), so accuracy alone can mislead; we recommend reporting macro-F1 alongside accuracy.

\section*{Ethics Statement}

The benchmark uses only publicly accessible OpenReview submissions, reviews, and discussions from NeurIPS~2024 and ICLR~2024; no private or proprietary data are used, and no reviewer or author is identified beyond what is already public. The 150 \textsc{Human-Verified} instances were labeled by two graduate-level researchers following explicit guidelines after a calibration round; no crowdsourced or low-paid labor was involved. \system\ is intended as an \emph{assistive} tool that supplements rather than replaces human judgment, and should not be used to rank or target specific reviewers, authors, or submissions. Because LLM-based judges and extractors can carry systematic biases (Section~\ref{sec:full_context_resuls}), automated labels should not be treated as ground truth without further validation. We release all data, labels, prompts, and code under open licenses with documentation of provenance and known limitations.

\bibliography{sample-base}
\newpage
\appendix
\clearpage

\section*{Appendix}


\paragraph{Appendix Overview.}
The appendix is organized around the three questions a reader is most likely to bring to the paper.
\emph{How was the benchmark built, and can its labels be trusted?} Appendix~\ref{app:construction} covers the construction pipeline end to end: the extractor comparison and why we run an open-weight model despite lower F1 (\ref{app:extraction}, \ref{app:extract-cost}), the human audit of the semantic-equivalence judge (\ref{app:judge-human}), label definitions and the factual/subjective split (\ref{app:label-examples}, \ref{app:factual}, \ref{app:claimtype}), and the two reliability analyses that matter most, agreement between the automated labels and human consensus (\ref{app:human-agreement}) and the two-labeler arbitration protocol behind the canonical labels (\ref{app:disagreement}).
\emph{What exactly was the model asked?} Appendix~\ref{app:prompts-grp} reproduces every prompt verbatim.
\emph{What do the numbers look like beyond the main tables?} Appendix~\ref{app:results-grp} holds the per-model results (\ref{app:rq1-table}, \ref{app:macro-f1}, \ref{app:b1-rag}, \ref{app:rag-extended}), the non-LLM entailment baselines (\ref{app:nli}), the manual error taxonomy and the calibration effect behind the \textsc{PaperSourced} ordering (\ref{app:error-analysis}, \ref{app:ps-preddist}), benchmark size in context (\ref{app:size}), retrieval quality and cost (\ref{app:retrieval}, \ref{app:token-efficiency}), and the reason intrinsic retrieval metrics apply to only one split (\ref{app:retrieval-scope}).
Appendix~\ref{app:related} gives the extended related work, and Appendix~\ref{app:deployment} closes with the intended deployment setting.

\section{Benchmark Construction and Annotation}
\label{app:construction}

\subsection{Claim Extraction Evaluation}
\label{app:extraction}

A critical step in the \system pipeline is accurately decomposing dense, multi-faceted reviewer comments into isolated and verifiable statements. Because reviewer feedback often blends factual assertions with subjective interpretation, the claim extraction module must distill these paragraphs into atomic claims while strictly preserving the original intent of the reviewer.
To determine the most robust approach for this task, table~\ref{tab:b2-unified} compares the performance of three distinct claim extractors (Fenice~\citep{scire2024fenice}, Gemma~\citep{hosseini2024scalable}, and \texttt{Qwen3-4B}~\citep{yang2025qwen3}) across two evaluation axes: semantic alignment with a reference extraction (LLM-as-a-judge) and intrinsic structural quality (reference-free). The two axes are complementary: reference-based metrics capture coverage, while reference-free metrics capture whether individual claims are verification-ready.
The evaluation focuses on how effectively each model parses authentic review text from our NeurIPS and ICLR 2024 dataset into self-contained propositions suitable for downstream retrieval and verification.

\subsubsection{Reference-Based Evaluation.}
To establish a strong comparison signal, we use \texttt{o4-mini} as a high-quality reasoning model to generate a reference set of claims that are manually verified to be atomic and decontextualized. We treat this reference extraction as a stronger summarization-style signal and evaluate how well each candidate extraction method aligns with it. Agreement is measured using an LLM-as-a-judge protocol. We employ \texttt{GPT-5-mini} as a semantic judge: for each extracted claim, we identify the most similar reference claim and ask the judge whether the two express the same atomic proposition. Based on these semantic equivalence decisions, we compute precision, recall, and F1 scores. 

For the reference-based axis, \texttt{o4-mini} is used \emph{only} to generate the high-fidelity reference set against which the three candidate extractors are scored; it is not used to produce any of the 800 benchmark claims. The benchmark claims are sourced exclusively from \texttt{Qwen3-4B}, which achieves the best overall balance and is therefore used for all benchmark construction, ensuring extraction quality and reproducibility.

\subsubsection{Reference-Free Evaluation.}
Regardless of the reference set, a useful claim must meet specific structural and linguistic criteria for downstream verification~\citep{ullrich2025claim, wright-etal-2022-generating}. We use \texttt{o4-mini} to audit each method's output against three core dimensions. \textbf{Atomicity} measures whether a claim expresses exactly one checkable assertion. \textbf{Faithfulness} assesses whether the claim remains grounded in the original reviewer's text without introducing hallucinations or scope-creep. \textbf{Decontextualization} checks whether the claim is self-contained and does not rely on unresolved references such as ``the method'' or ``Figure~1'' that require surrounding context to interpret.

\begin{table}[t]
\centering

\caption{Claim-extraction quality: LLM-as-a-judge (\texttt{GPT-5-mini}, reference-based) and reference-free intrinsic metrics.}
\resizebox{\linewidth}{!}{  
\begin{tabular}{l cccc ccc}
\toprule
 &&
\multicolumn{3}{c}{\textbf{Reference-Based}} &
\multicolumn{3}{c}{\textbf{Reference-Free}} \\
\cmidrule(lr){3-5}
\cmidrule(lr){6-8}
\textbf{Extractor}  & \textbf{count} &
\textbf{Prec.} & \textbf{Rec.} & \textbf{F1} &
\textbf{Atom.} & \textbf{Flue.} & \textbf{Deco.} \\
\midrule
Fenice  & 739 &
0.872 & 0.538 & 0.665 &
0.314 & 0.788 & 0.362 \\

Gemma  & 2464 &
0.868 & \textbf{0.762} & \textbf{0.811} &
0.379 & 0.781 & 0.329 \\

Qwen3-4b  & 923 &
\textbf{0.889} & 0.587 & 0.707 &
\textbf{0.451} & \textbf{0.927} & \textbf{0.524} \\
\bottomrule
\end{tabular}
}
\label{tab:b2-unified}
\end{table}

\bigskip

\paragraph{Gemma extracts the most claims; \texttt{Qwen3-4B} extracts the most usable ones.}

The evaluation reveals distinct operational profiles among the extracted models. Fenice exhibits a high-precision, low-recall regime by extracting 739 claims with a precision of 0.872 and a recall of 0.538, effectively minimizing false positives at the expense of comprehensive coverage. Conversely, Gemma prioritizes recall by achieving a rate of 76.2\% and an F1 score of 0.811, but it generates a substantially larger volume of 2,464 candidate claims. \texttt{Qwen3-4B} optimizes this balance by producing 923 claims with a precision of 0.889, thereby maintaining strict semantic control. In the context of downstream verification, the implications of this tradeoff are asymmetric: while unextracted claims merely reduce total evaluation coverage, poorly formulated claims systematically confound reliable classification.

\paragraph{Intrinsic quality is the discriminating dimension.}

Beyond semantic alignment, intrinsic structural quality emerges as the critical factor determining extractor viability. \texttt{Qwen3-4B} demonstrates significantly superior performance across essential structural metrics. Specifically, it achieves an atomicity score of 0.451, which exceeds the scores of 0.379 for Gemma and 0.314 for Fenice. Furthermore, its fluency score of 0.927 surpasses both Gemma and Fenice as well. Finally, \texttt{Qwen3-4B} records a decontextualization score of 0.524, notably outperforming the respective scores of 0.329 and 0.362 achieved by Gemma and Fenice. These structural properties are foundational to downstream verification accuracy. Consequently, the high-fidelity benchmark slices, specifically \textsc{Verifiable-Only} and \textsc{HighAgreement}, are constructed exclusively using \texttt{Qwen3-4B} extractions. Robust claim extraction is a prerequisite for reliable evaluation, meaning these extraction quality differentials directly determine the diagnostic validity of the resulting benchmark partitions.

\subsection{Extraction Cost, Latency, and Model Choice}
\label{app:extract-cost}

Table~\ref{tab:b2-unified} compares the three open-weight extractors on quality alone. That comparison leaves an obvious question open, since \texttt{o4-mini} and \texttt{GPT-5-mini} already appear elsewhere in the pipeline: if those models are good enough to generate the reference claims and to judge semantic equivalence, why not use one of them to extract? Table~\ref{tab:extract-cost} answers it with measurements instead of an assertion, scoring all five candidates under the same protocol and adding median latency per review and average API cost per review.

\begin{table}[t]
\centering
\caption{Claim-extraction quality, latency, and cost for all five candidate extractors, scored against the same reference claim set with the \texttt{GPT-5-mini} semantic-equivalence judge. Latency is the median wall-clock time per review, measured locally on a single NVIDIA RTX~3090 for open-weight models and over the network for API models. Cost is the average API spend per review. $^{\dagger}$\texttt{GPT-5-mini} also generated the reference claims, so its F1 is a self-consistency upper bound rather than a comparable score; \texttt{o4-mini} is the fair proprietary reference point.}
\resizebox{\columnwidth}{!}{%
\begin{tabular}{lccccc}
\toprule
\textbf{Extractor} & \textbf{Prec.} & \textbf{Rec.} & \textbf{F1} & \textbf{Latency (s)} & \textbf{Cost} \\
\midrule
Fenice              & 0.872 & 0.538 & 0.665 & \phantom{0}1.97 & open-weight \\
\texttt{Qwen3-4B}   & 0.889 & 0.587 & 0.707 & \phantom{0}8.20 & open-weight \\
Gemma               & 0.868 & 0.762 & 0.811 & 21.76 & open-weight \\
\midrule
\texttt{o4-mini}    & 0.990 & 0.817 & 0.895 & \phantom{0}9.62 & \$0.29 \\
\texttt{GPT-5-mini} & \textbf{0.994} & \textbf{0.849} & 0.915$^{\dagger}$ & 24.43 & \$0.11 \\
\bottomrule
\end{tabular}%
}
\label{tab:extract-cost}
\end{table}

The proprietary models are clearly better extractors. \texttt{o4-mini} reaches 0.895 F1 and \texttt{GPT-5-mini} 0.915, against 0.707 for \texttt{Qwen3-4B}. We note that the \texttt{GPT-5-mini} figure is not directly comparable, because the same model generated the reference claims, so 0.915 is a self-consistency ceiling; \texttt{o4-mini} at 0.895 is the honest cross-model number.

We still extract with \texttt{Qwen3-4B}, for three reasons that the table makes concrete. First, cost scales with volume, and extraction is the highest-volume stage: it runs over every sentence of every review, whereas verification runs once per extracted claim. At \$0.11 to \$0.29 per review, extracting a full conference cycle through an API is a real budget line and a rate-limit exposure, and it puts the expensive model at the cheap end of the pipeline. We would rather spend that budget on verification, which is where the reasoning difficulty actually is. Second, reproducibility: an open-weight extractor lets anyone regenerate or extend the benchmark, whereas a versioned API can be updated or deprecated underneath a published dataset. Third, the quality gap sits mostly in recall (0.587 against 0.817), not precision (0.889 against 0.990). For a benchmark that evaluates claim \emph{verification} rather than exhaustive claim \emph{mining}, missing a claim reduces coverage while a badly formed claim corrupts a verification unit, so precision is the axis that matters and the local model is close to the frontier there.

\subsection{Human Validation of the Semantic-Equivalence Judge}
\label{app:judge-human}

The reference-based scores in Table~\ref{tab:b2-unified} rest on \texttt{GPT-5-mini} deciding whether an extracted claim and a reference claim express the same proposition. Since that decision drives which extractor we adopt, we validate it against human annotation like every other automated component of the benchmark. An annotator who had not worked on the extraction pipeline independently re-decided 50 randomly sampled equivalence judgements per extractor, 150 in total, seeing the claim pair but not the judge's verdict.

\begin{table}[t]
\centering
\caption{Human validation of the \texttt{GPT-5-mini} semantic-equivalence judge used for the reference-based extraction scores. An annotator who was not involved in the automated pipeline independently re-decided 50 randomly sampled equivalence judgements per extractor.}
\resizebox{0.9\columnwidth}{!}{%
\begin{tabular}{lcc}
\toprule
\textbf{Extractor} & \textbf{Agreement} & \textbf{Rate} \\
\midrule
Fenice            & 45 / 50  & 90.0\% \\
Gemma             & 42 / 50  & 84.0\% \\
\texttt{Qwen3-4B} & 47 / 50  & \textbf{94.0\%} \\
\midrule
\textbf{Overall}  & \textbf{134 / 150} & \textbf{89.3\%} \\
\bottomrule
\end{tabular}%
}
\label{tab:judge-human}
\end{table}

Agreement is 89.3\% overall and is consistent across extractors, from 84.0\% on Gemma to 94.0\% on \texttt{Qwen3-4B} (Table~\ref{tab:judge-human}). This is close to the rate at which our groundedness labels match human consensus (Appendix~\ref{app:human-agreement}), which suggests the judge is a reasonable stand-in for a human on this narrow decision, and it supports applying it across the whole extraction evaluation rather than reporting a hand-checked subset. The Gemma figure is the lowest of the three, which is consistent with its output profile: it produces 2{,}464 claims, many of them long or under-decontextualized, and those are exactly the pairs where equivalence is a judgement call rather than a lookup.

\subsection{Label Definitions and Examples}
\label{app:label-examples}

Each review claim receives one of the four labels defined in Section~\ref{sec:task-formulation}, determined by whether the manuscript provides sufficient evidence under a strict document-bounded grounding criterion. Worked examples follow.
\begin{itemize}[leftmargin=1.5em, itemsep=2pt, topsep=2pt]
  \item \texttt{Supported}: The manuscript provides clear and sufficient evidence that fully substantiates the claim. \emph{Example:} A reviewer states ``The authors evaluate on CIFAR-100,'' and the paper explicitly reports CIFAR-100 experiments.
  \item \texttt{Not Supported}: The manuscript contradicts the claim or the asserted content is demonstrably absent. \emph{Example:} A reviewer states ``No ablation study is provided,'' but the paper contains a dedicated ablation section.
  \item \texttt{Partially Supported}: The manuscript supports only a qualified or incomplete version of the claim. \emph{Example:} A reviewer states ``The method outperforms all baselines,'' but the paper shows it outperforms most but not all.
  \item \texttt{Not Determined}: The manuscript does not contain sufficient information to resolve the claim in either direction. This label applies when (i)~the claim references external knowledge or prior work not discussed in the paper, (ii)~the claim is too vague or ambiguous to verify, or (iii)~the relevant evidence is absent without any contradicting assertion.
\end{itemize}


\subsection{Factual vs.\ Subjective Claims}
\label{app:factual}

Each claim is tagged as \emph{factual} or \emph{subjective} using \texttt{o4-mini}: \emph{factual} if its validity is grounded in content observable in the manuscript (e.g., a missing experiment or an absent figure), and \emph{subjective} if it is an opinion, recommendation, or forward-looking judgment about novelty, significance, or impact that cannot be resolved from the paper alone. The \textsc{verifiable-only} slice retains only factual claims, so verification on that slice targets assertions checkable against the manuscript rather than reviewer opinion.

\subsection{Claim Types and Arbitration}
\label{app:claimtype}

We used \texttt{GPT-5-mini} to sort the 800 benchmark claims into objective claims, whose truth follows from something written in the manuscript, and subjective claims, which are judgements about novelty, significance, or direction. The split is 444 objective and 356 subjective. We then measured how often each type required human arbitration, meaning the two automated labelers disagreed and a person had to break the tie.

\begin{table}[t]
\centering
\caption{Arbitration on \textsc{RebuttalSourced} broken down by claim type. \emph{Arbitrated} counts claims where the two automated labelers disagreed and a human tie-breaker was needed. \emph{Neither} counts disagreements where the human annotator supplied a third label because both candidates were wrong.}
\resizebox{\columnwidth}{!}{%
\begin{tabular}{lccc}
\toprule
\textbf{Claim type} & \textbf{Total} & \textbf{Arbitrated} & \textbf{Arbitration rate} \\
\midrule
Objective  & 444 & 184 & 41.4\% \\
Subjective & 356 & 136 & 38.2\% \\
\midrule
Overall    & 800 & 320 & 40.0\% \\
\bottomrule
\end{tabular}%
}

\vspace{0.6em}

\resizebox{\columnwidth}{!}{%
\begin{tabular}{lcc}
\toprule
\textbf{Arbitration outcome} & \textbf{Objective} & \textbf{Subjective} \\
\midrule
Labelers agree (no arbitration) & 260 (58.6\%) & 220 (61.8\%) \\
\texttt{GPT-5-mini} upheld      & \phantom{0}93 (21.0\%) & \phantom{0}65 (18.3\%) \\
\texttt{Claude-Sonnet-4-6} upheld & \phantom{0}79 (17.8\%) & \phantom{0}57 (16.0\%) \\
Neither upheld                  & \phantom{0}12 (2.7\%)  & \phantom{0}14 (3.9\%) \\
\bottomrule
\end{tabular}%
}
\label{tab:claimtype-arbitration}
\end{table}

The result runs against intuition. Objective claims are arbitrated \emph{more} often than subjective ones, 41.4\% against 38.2\% (Table~\ref{tab:claimtype-arbitration}). The explanation is that objectivity in a peer review is not the same as ease of verification. Settling ``the paper does not compare against baseline X'' or ``the reported IoU is 44.91'' requires finding one specific fact somewhere in a long manuscript, and two labelers that retrieve different passages will reach different conclusions. Subjective claims are vaguer, but that vagueness gives the labelers less to disagree about: both tend to land on \texttt{Partially Supported} or \texttt{Not Determined}.

The arbitration outcomes point the same way. \texttt{GPT-5-mini} is upheld somewhat more often than \texttt{Claude-Sonnet-4-6} in both categories (21.0\% against 17.8\% on objective claims, 18.3\% against 16.0\% on subjective ones), so neither labeler dominates and the arbitration is not a systematic correction of one model. Cases where the human annotator had to supply a third label because both candidates were wrong are rare, 2.7\% on objective claims and 3.9\% on subjective ones. The slightly higher rate on subjective claims is the expected direction: when a reviewer's remark is genuinely interpretive, there is sometimes no label either model would have proposed.

\subsection{Benchmark Label Distributions}
\label{app:distributions}

Table~\ref{tab:peercheck-distributions} in Section~\ref{sec:dataset} reports the full \system benchmark statistics overview, including label distributions across all six benchmark variants and dataset provenance (venue and acceptance decision). The \textsc{PaperSourced} variant is all-Supported by construction; the remaining variants show progressively more heterogeneous label distributions as claims become more interaction-grounded and ambiguous. The four \textsc{Human-Verified} label counts are the human-consensus totals of Table~\ref{tab:hv-confusion}.

\subsection{Inter-Source Agreement}
\label{app:agreement}

The \textsc{HighAgreement} slice is constructed by retaining only instances where \textsc{RebuttalSourced} and \textsc{LLMJudged} labels agree. This section describes the agreement analysis underlying this design choice.

\paragraph{Automated supervision sources.}
\textsc{RebuttalSourced} labels are derived from author--reviewer interaction dynamics: the model infers whether the author's response confirms or refutes the reviewer's claim. \textsc{LLMJudged} labels are derived from direct document-bounded verification: the model checks whether the manuscript supports the claim without consulting the discussion thread. These two sources are therefore \emph{independent} in both mechanism and evidence source, making their agreement a meaningful signal of claim clarity and label reliability.

Across the 800 core review-derived claims, the canonical \textsc{RebuttalSourced} and \textsc{LLMJudged} labels agree on 175 instances (21.9\%), forming the \textsc{HighAgreement} slice. The relatively low raw agreement rate reflects the genuine difficulty of peer-review claims and the different perspectives captured by the two supervision sources: interaction-grounded labels capture how authors interpret their own work, while document-grounded labels capture what the manuscript literally supports.

\paragraph{Automated supervision vs.\ human labels.}
To validate automated supervision quality, we compare both \textsc{RebuttalSourced} and \textsc{LLMJudged} labels against the \textsc{Human-Verified} consensus labels. Inter-annotator agreement between the two human annotators is near-perfect (Cohen's $\kappa \approx 0.96$), indicating that the high human--automated agreement is not an artifact of low human reliability. These results support the use of \textsc{RebuttalSourced} and \textsc{LLMJudged} as reliable large-scale supervision sources, with \textsc{Human-Verified} serving as a gold-standard reference for validation and error analysis. Appendix~\ref{app:human-agreement} gives the full confusion matrix and a per-claim-type breakdown.

\subsection{Human-Verified Label Quality}
\label{app:human-agreement}

Most of the benchmark is labeled automatically, so the question that matters is not whether those labels look plausible but how far they can be trusted. The \textsc{Human-Verified} subset exists to answer it: 300 claims, 37.5\% of the benchmark, labeled by hand under the guidelines of Section~\ref{sec:humanjudge} and used as the reference against which the automated supervision is scored.

\begin{table}[t]
\centering
\caption{Confusion matrix between automated (\textsc{LLMJudged}) labels and human consensus labels on the expanded \textsc{Human-Verified} subset ($n=300$). Rows are automated labels, columns are human labels. Exact four-way agreement is the diagonal, $271/300 = 90.3\%$ (Cohen's $\kappa = 0.87$).}
\resizebox{\columnwidth}{!}{%
\begin{tabular}{lcccc}
\toprule
 & \multicolumn{4}{c}{\textbf{Human}} \\
\cmidrule(lr){2-5}
\textbf{Automated} & \textbf{Supp.} & \textbf{Part.\ Supp.} & \textbf{Not Supp.} & \textbf{Not Det.} \\
\midrule
\texttt{Supported}           & \textbf{65} & 2   & 1  & 1  \\
\texttt{Partially Supported} & 8  & \textbf{102} & 7  & 0  \\
\texttt{Not Supported}       & 2  & 3   & \textbf{69} & 2  \\
\texttt{Not Determined}      & 2  & 0   & 1  & \textbf{35} \\
\bottomrule
\end{tabular}%
}
\label{tab:hv-confusion}
\end{table}

\paragraph{Automated labels match human consensus on 90.3\% of claims ($\kappa = 0.87$).}
Under a strict four-way exact-match criterion, the automated labels agree with human consensus on 271 of 300 claims, 90.3\%, with Cohen's $\kappa = 0.87$ \citep{cohen1960kappa}, which falls in the ``almost perfect'' band on the conventional scale \citep{landis1977measurement}. Table~\ref{tab:hv-confusion} shows where the 29 disagreements sit, and their shape matters more than their count. They cluster on the \texttt{Partially Supported} boundary: 8 claims the humans called \texttt{Supported} were labeled \texttt{Partially Supported}, and 7 they called \texttt{Not Supported} were labeled the same way. Under an ordinal-tolerant criterion, where a one-step difference on the \texttt{Supported}, \texttt{Partially Supported}, \texttt{Not Supported} scale is acceptable, agreement rises to 97.0\%. Hard polarity flips, where one side says clearly supported and the other says clearly unsupported, occur 3 times in 300, or 1.0\%. The automated supervision is therefore not making a different kind of judgement from the humans; it draws the partial-support line in a slightly different place, which is the same place two human annotators need a consensus discussion.

\begin{table}[t]
\centering
\caption{Agreement between automated labels and human consensus on the expanded \textsc{Human-Verified} subset, broken down by claim type. Reliability is essentially unchanged between objective and subjective claims, which is the case an annotation-bias account would predict to diverge.}
\resizebox{\columnwidth}{!}{%
\begin{tabular}{lccc}
\toprule
\textbf{Slice} & \textbf{n} & \textbf{Agreement} & \textbf{Cohen's $\kappa$} \\
\midrule
Overall                & 300 & 90.3\% & 0.866 \\
Objective / factual    & 158 & 90.5\% & 0.866 \\
Subjective             & 142 & 90.1\% & 0.858 \\
\bottomrule
\end{tabular}%
}
\label{tab:hv-agreement}
\end{table}

\paragraph{Reliability is the same on objective and subjective claims.}
A bias inherited from the labeler models would most plausibly show up as degraded reliability on exactly the claims where reviewer language is interpretive. It does not. Table~\ref{tab:hv-agreement} splits the subset by claim type: 90.5\% agreement and $\kappa = 0.866$ on the 158 objective claims, 90.1\% and $\kappa = 0.858$ on the 142 subjective ones. A 0.4-point difference at this sample size is not a signal. The automated labels track human judgement about equally well on both, which is the evidence we can offer that they reflect claim groundedness rather than an artifact of the annotation procedure. It does not prove the absence of a bias that the models and the guidelines share, and we say so in the Limitations section.

\begin{table}[t]
\centering
\caption{RAG accuracy on the expanded \textsc{Human-Verified} subset ($n=300$), by retriever. These are the values reported in the \textsc{Human-Verified} column of Table~\ref{tab:rag-b2-b6}. No model clears $0.35$ under any retriever, and swapping retrievers moves accuracy by at most a few points, which is the pattern we use to argue that reasoning rather than retrieval is the binding constraint on this split.}
\resizebox{\columnwidth}{!}{%
\begin{tabular}{lcccc}
\toprule
\textbf{Model} & \textbf{BM25} & \textbf{BM25+R} & \textbf{Dense} & \textbf{Dense+R} \\
\midrule
\texttt{o4-mini}      & 0.243 & 0.237 & 0.230 & 0.257 \\
\texttt{GPT-5-mini}   & 0.273 & 0.290 & 0.297 & 0.257 \\
\texttt{Claude-Haiku} & 0.270 & 0.240 & 0.237 & 0.300 \\
\texttt{Qwen3-8B}     & 0.257 & 0.297 & 0.283 & 0.277 \\
\texttt{Qwen2.5-7B}   & 0.300 & \textbf{0.340} & 0.307 & 0.287 \\
\bottomrule
\end{tabular}%
}
\label{tab:hv-rag-300}
\end{table}

\paragraph{No verifier clears 0.38 on the hand-labeled claims.}
\textsc{Human-Verified} is among the hardest splits in the benchmark. No verifier clears 0.38 in full-context (Table~\ref{tab:rq1-fullpaper}) or 0.34 under any retriever (Table~\ref{tab:hv-rag-300}). \texttt{Qwen3-8B} is the strongest full-context model here at 0.377, which is consistent with the subset being dominated by \texttt{Partially Supported} claims (107 of 300) and with that model's tendency to over-predict exactly that label (Figure~\ref{fig:pred-dist}). Because the human labels are the reference rather than a proxy, these numbers are the cleanest available estimate of how far current verifiers are from the task.

\subsection{Annotator Disagreement and Arbitration}
\label{app:disagreement}

The canonical labels for \textsc{RebuttalSourced} and \textsc{LLMJudged} are produced by a two-annotator protocol with human arbitration. Each of the 800 atomic reviewer claims is labeled twice and independently: \texttt{GPT-5-mini} (Annotator~A) and \texttt{Claude-Sonnet-4-6} (Annotator~B) receive the same claim and the same source material (the author response for \textsc{RebuttalSourced} and the paper content for \textsc{LLMJudged}) under identical prompts. For every record where the two annotators disagree, a human annotator is shown the claim, the relevant source material, and both candidate labels (with model identity anonymized and presentation order randomized per record), and either selects the correct label or assigns a new one when both candidates are wrong. This arbitrated label is taken as canonical.

\paragraph{Inter-annotator agreement before arbitration.}
Table~\ref{tab:interannotator} reports raw agreement between the two annotators. \texttt{Claude-Sonnet-4-6} is systematically more conservative than \texttt{GPT-5-mini}: on \textsc{RebuttalSourced} it reclassifies many \texttt{Supported} calls as \texttt{Partially Supported} or \texttt{Not Supported}, and on \textsc{LLMJudged} it shifts confident labels into \texttt{Not Determined}. The two models agree almost entirely on \texttt{Not Determined} itself (52/67 on \textsc{RebuttalSourced}, 193/251 on \textsc{LLMJudged}) but diverge on the confident labels.

\begin{table}[t]
\centering
\caption{Inter-annotator agreement (\texttt{GPT-5-mini} vs.\ \texttt{Claude-Sonnet-4-6}) before arbitration.}
\label{tab:interannotator}
\resizebox{\columnwidth}{!}{%
\begin{tabular}{lccc}
\toprule
\textbf{Benchmark} & \textbf{n} & \textbf{Annotators agree} & \textbf{Rate} \\
\midrule
\textsc{RebuttalSourced} & 800 & 480 & 60.0\% \\
\textsc{LLMJudged}       & 800 & 459 & 57.4\% \\
\bottomrule
\end{tabular}%
}
\end{table}

\paragraph{Arbitration verdicts.}
Table~\ref{tab:arbitration} shows, among the disagreements, how often each annotator was upheld by the human annotator. \texttt{GPT-5-mini} is correct slightly more often ($\sim$49\%) than \texttt{Claude-Sonnet-4-6} ($\sim$41\%), but in roughly one in ten disagreements neither candidate was correct and the annotator supplied a third label.

\begin{table}[t]
\centering
\caption{Quantitative breakdown of arbitration verdicts on annotator disagreements. Annotator A and B denote the \texttt{GPT-5-mini} and \texttt{Claude-Sonnet-4-6}, respectively.}
\label{tab:arbitration}
\resizebox{\columnwidth}{!}{%
\begin{tabular}{lcccc}
\toprule
\textbf{Benchmark} & \textbf{\#Disagree} & \textbf{A correct} & \textbf{B correct} & \textbf{Neither} \\
\midrule
\textsc{RebuttalSourced} & 320 & 158 (49.4\%) & 136 (42.5\%) & 26 (8.1\%) \\
\textsc{LLMJudged}       & 341 & 168 (49.3\%) & 137 (40.2\%) & 35 (10.3\%) \\
\bottomrule
\end{tabular}%
}
\end{table}

\paragraph{Canonical distribution and derived slices.}
Table~\ref{tab:canonical} gives the arbitrated label distribution. The \textsc{LLMJudged} distribution is heavily weighted toward \texttt{Not Determined}, reflecting that a meaningful share of reviewer claims have no verifiable answer in the paper text alone. After arbitration, the \textsc{HighAgreement} slice (canonical \textsc{RebuttalSourced} label equals canonical \textsc{LLMJudged} label) contains 175 instances, and the \textsc{verifiable-only} slice contains 131.

\begin{table}[t]
\centering
\caption{Canonical label distribution (post-arbitration).}
\label{tab:canonical}
\resizebox{\columnwidth}{!}{%
\begin{tabular}{lcc}
\toprule
\textbf{Label} & \textbf{\textsc{RebuttalSourced}} & \textbf{\textsc{LLMJudged}} \\
\midrule
\texttt{Supported}           & 217 & 207 \\
\texttt{Partially Supported} & 310 & 93 \\
\texttt{Not Supported}       & 189 & 141 \\
\texttt{Not Determined}      & 84  & 359 \\
\bottomrule
\end{tabular}
}
\end{table}

\section{Related Works}
\label{app:related}

\paragraph{LLMs for Peer Review.}
A growing body of work explores the use of large language models to support scholarly peer review through tasks such as review generation, review assistance, reviewer recommendation, and manuscript assessment \citep{yuan2022reviewgen,liang2024feedback,tomkins2017reviewer}. A related line analyzes peer-review content and dynamics, including review quality, argumentation, sentiment, and reviewer behavior \citep{hua2019argument,rogers2020peerreview,kang2018dataset,ebrahimi2025rottenreviews}. These efforts demonstrate that state-of-the-art language models can generate meaningful feedback and shed light on how reviews are written and how they function socially. However, their primary objective is to produce, evaluate, or characterize reviews rather than to determine whether specific claims made within a review are supported by evidence contained in the reviewed manuscript. As a result, the problem of manuscript-grounded review verification remains largely unexplored.

\paragraph{Scientific Claim Verification.}
\system\ is also related to scientific claim verification and evidence-based fact checking, studied extensively through textual entailment and natural language inference, from early RTE-style formulations \citep{dagan2006rte} to large-scale benchmarks such as FEVER \citep{thorne2018fever}, SciFact \citep{wadden2020scifact}, PubHealth \citep{kotonya2020explainable}, and scientific verification tasks \citep{malon2018ci,wadden2021sciver}. These datasets have enabled systems that retrieve evidence and determine whether a claim is supported, contradicted, or unsupported, and retrieval-augmented approaches \citep{lewis2020rag,izacard2021distilling} have further improved evidence-grounded and multi-hop reasoning. While these settings share similarities with review verification, they typically treat claims as standalone statements and assume open-domain retrieval or citation-based reasoning across multiple sources, with evaluation centered on short claims whose supporting evidence is relatively concentrated. In contrast, peer-review claims often contain methodological observations, novelty assessments, and experimental critiques whose supporting evidence may be distributed across multiple sections of a single manuscript. Review verification is therefore a document-bounded grounding problem that requires reasoning over long, highly structured scientific documents while accounting for contextual qualifiers, experimental assumptions, and evidence aggregation.

\paragraph{Claim Decomposition.}
A further challenge arises from the fact that review comments frequently contain multiple assertions expressed within a single sentence or paragraph. Recent work on claim decomposition has shown that breaking complex statements into atomic claims improves the reliability and interpretability of downstream verification \citep{metropolitansky2025claimify,min2023factscore,scire2024fenice}. Peer reviews complicate this step because they often mix subjective assessment with factual assertions and rely on hedging or implicit references. \system\ adopts a similar atomic-claim perspective and adapts extraction using multiple supervision signals, enabling more precise evidence attribution and systematic, manuscript-grounded groundedness assessment in an end-to-end evaluation.

\paragraph{Concurrent Work.}
A closely related concurrent system is FactReview~\citep{xu2026factreview}, which also targets evidence-grounded peer review assessment. FactReview focuses on author-written claims extracted from the submitted manuscript, using literature retrieval and code execution to verify whether reported results are reproducible and positioned correctly relative to prior work. \system\ is complementary in scope and emphasis: we focus on \emph{reviewer}-authored claims, statements made in the review about the paper rather than by the paper, and derive supervision from author--reviewer interaction dynamics and multiple independent label sources. This distinction matters because reviewer claims introduce interpretive and normative language absent from author-written text, and their groundedness must be assessed against a single manuscript rather than the broader literature. Our multi-benchmark design, spanning six supervision variants with different reliability and ambiguity profiles, enables systematic evaluation of the task difficulty spectrum.

\clearpage
\section{Prompts}
\label{app:prompts-grp}

This section lists every prompt used in the pipeline verbatim. Figure~\ref{fig:prompt-verify} gives the groundedness verification prompt shared by all five verifiers, Figure~\ref{fig:prompt-extract} the claim-extraction prompt, and Figure~\ref{fig:prompt-label} the two labeling prompts behind \textsc{RebuttalSourced} and \textsc{LLMJudged}.

\subsection{Verification Prompt Template}
\label{app:prompt}

We use the following prompt template for all LLM verifiers in full-context mode. The same label definitions are used across all dataset variants. For RAG-based verification, the \texttt{\{manuscript\}} field is replaced with the concatenation of top-$k$ retrieved passages, ranked by the chosen retrieval configuration.

\begin{figure*}[t]
\begin{tcolorbox}[promptbox={promptVerify}{Groundedness Verification Prompt (full-context and RAG)}]
You are a scientific claim verifier. Given a peer review claim and the full text of the reviewed manuscript, determine whether the manuscript supports the claim.

\medskip
\textbf{Claim:} \texttt{\{claim\}}\qquad\textbf{Manuscript:} \texttt{\{manuscript\}}

\medskip
Classify the claim as exactly one of the following labels based solely on the content of the manuscript:

\begin{itemize}[leftmargin=1.5em, itemsep=2pt, topsep=2pt]
  \item \textbf{Supported:} The manuscript provides clear and sufficient evidence that fully substantiates the claim.
  \item \textbf{Not Supported:} The manuscript contradicts the claim, or the asserted content is demonstrably absent from the manuscript.
  \item \textbf{Partially Supported:} The manuscript supports only a qualified or incomplete version of the claim (e.g., missing conditions, limited scope, or omitted caveats).
  \item \textbf{Not Determined:} The manuscript does not contain sufficient information to resolve the claim in either direction. Use this label when (1) the claim references external knowledge, (2) the claim is too vague or ambiguous to verify, or (3) the relevant evidence is absent without any contradicting assertion.
\end{itemize}
\end{tcolorbox}
\caption{Groundedness verification prompt, shared by all five verifiers. Under RAG the \texttt{\{manuscript\}} field carries the top-$k$ retrieved passages instead of the full paper.}
\label{fig:prompt-verify}
\end{figure*}

\subsection{Claim Extraction Prompt}
\label{app:prompt-extract}

The prompted extractor (\texttt{Qwen3-4B}) decomposes each review into atomic claims with the prompt in Figure~\ref{fig:prompt-extract}.

\begin{figure*}[t]
\begin{tcolorbox}[promptbox={promptExtract}{Claim Extraction Prompt (\texttt{Qwen3-4B})}]
\textbf{System:} You are an expert at extracting factual claims from academic reviews.

\medskip
\textbf{User:} Extract factual claims from the review text below. A claim is a specific, verifiable statement about the paper.

\textbf{Review Text:} \texttt{\{review\_text\}}

\medskip
Each claim should be (1) a single atomic statement, (2) verifiable against the paper or author response, and (3) specific and factual (not an opinion).
\end{tcolorbox}
\caption{Claim-extraction prompt used to decompose each review comment into atomic, self-contained claims.}
\label{fig:prompt-extract}
\end{figure*}

\begin{table}[t]
\centering
\caption{\textsc{PaperSourced} Dense+R accuracy vs.\ the Oracle upper bound (gold paragraph fed directly) for all five models. Full four-retriever results in Table~\ref{tab:rag-b2-b6}.}
\resizebox{\columnwidth}{!}{%
\begin{tabular}{lcc}
\toprule
\textbf{Model} & \textbf{Dense+Reranker} & \textbf{Oracle} \\
\midrule
\texttt{o4-mini}          & \textbf{0.904} & \textbf{0.988} \\
\texttt{GPT-5-mini}       & 0.852          & 0.958          \\
\texttt{Claude-Haiku}     & 0.844          & 0.860          \\
\texttt{Qwen3-8B}         & 0.814          & 0.872          \\
\texttt{Qwen2.5-7B}       & 0.860          & 0.816          \\
\bottomrule
\end{tabular}
}
\label{tab:rq2-b1}
\end{table}

\subsection{Labeling Prompts}
\label{app:prompt-label}

The two automated annotators (\texttt{GPT-5-mini} and \texttt{Claude-Sonnet-4-6}) use the prompts in Figure~\ref{fig:prompt-label} for the \textsc{RebuttalSourced} and \textsc{LLMJudged} variants. The raw \texttt{Contradicted} label corresponds to \textit{Not Supported} in the four-way scheme.

\begin{figure*}[t]
\begin{tcolorbox}[promptbox={promptLabel}{\textsc{RebuttalSourced} Labeling Prompt (evidence: the discussion thread)}]
\textbf{System:} You are an expert at analyzing academic discourse.

\medskip
\textbf{User:} Given a reviewer's claim and the author's response, determine how the authors address the claim.

\textbf{Guidelines:} \emph{Supported}: authors clearly agree with or confirm the claim; \emph{Partially Supported}: authors acknowledge some validity but not fully; \emph{Contradicted}: authors explicitly disagree or provide counter-evidence; \emph{Not Determined}: authors do not address the claim, or it is unclear.

\medskip
\textbf{Reviewer Claim:} \texttt{\{claim\}}\qquad\textbf{Author's Response:} \texttt{\{author\_response\}}
\end{tcolorbox}

\vspace{0.6em}

\begin{tcolorbox}[promptbox={promptLabel}{\textsc{LLMJudged} Labeling Prompt (evidence: the manuscript)}]
\textbf{System:} You are an expert at analyzing academic papers.

\medskip
\textbf{User:} Given a reviewer's claim about a paper and the paper content, determine whether the claim is true according to the paper.

\textbf{Guidelines:} \emph{Supported}: the claim is true according to the paper; \emph{Partially Supported}: partially true; \emph{Contradicted}: the paper contradicts the claim; \emph{Not Determined}: the paper does not address the topic, or there is insufficient information. For example, if the claim states ``the paper lacks comparison with baseline X'' and the paper indeed omits it, the label is \emph{Supported}.

\medskip
\textbf{Reviewer Claim:} \texttt{\{claim\}}\qquad\textbf{Paper Content:} \texttt{\{paper\_content\}}
\end{tcolorbox}
\caption{The two labeling prompts. They differ only in the evidence they see: \textsc{RebuttalSourced} reads the author--reviewer thread and never the manuscript, \textsc{LLMJudged} reads the manuscript and never the thread. This disjointness is what makes their agreement informative (Appendix~\ref{app:agreement}).}
\label{fig:prompt-label}
\end{figure*}

\section{Additional Results}
\label{app:results-grp}

\subsection{Full-Context Verification Results}
\label{app:rq1-table}

\begin{table*}[t]
\centering
\caption{Full-context accuracy (five models, six benchmarks). \texttt{Qwen2.5-7B} excludes 9 over-length papers (Section~\ref{sec:baselines}). \textsc{Human-Verified} is the expanded 300-instance subset; Table~\ref{tab:hv-150-vs-300} compares it against the original 150 instances.}
\resizebox{0.85\linewidth}{!}{
\begin{tabular}{lccccc}
\toprule
\textbf{Benchmark} & \texttt{o4-mini} & \texttt{GPT-5-mini} & \texttt{Claude-Haiku} & \texttt{Qwen3-8B} & \texttt{Qwen2.5-7B} \\
\midrule
\textsc{PaperSourced}    & \textbf{0.866} & 0.792 & 0.764 & 0.618 & 0.753 \\
\textsc{RebuttalSourced} & 0.245 & 0.280 & \textbf{0.295} & 0.276 & 0.278 \\
\textsc{LLMJudged}       & \textbf{0.442} & 0.379 & 0.295 & 0.171 & 0.170 \\
\textsc{HighAgreement}   & \textbf{0.536} & 0.513 & 0.445 & 0.302 & 0.213 \\
\textsc{verifiable-only} & 0.495 & \textbf{0.518} & 0.388 & 0.320 & 0.250 \\
\textsc{Human-Verified}  & 0.267 & 0.317 & 0.300 & \textbf{0.377} & 0.256 \\
\bottomrule
\end{tabular}}
\label{tab:rq1-fullpaper}
\end{table*}

Table~\ref{tab:rq1-fullpaper} reports the precise full-context verification accuracy for all five models across the six benchmarks, complementing Figure~\ref{fig:rq1-accuracy}.

\subsection{Accuracy vs.\ Macro-F1}
\label{app:macro-f1}

\begin{table*}[t]
\centering
\caption{Accuracy vs.\ macro-F1 (full-context FC and best-retriever RAG) on \textsc{RebuttalSourced} and \textsc{LLMJudged}. A large Acc$-$F1 gap reflects label-distribution bias, not reasoning. \texttt{Qwen2.5-7B} FC excludes 9 over-length papers.}

\resizebox{0.85\textwidth}{!}{%
\begin{tabular}{lcccc cccc}
\toprule
 & \multicolumn{4}{c}{\textbf{Full-Context}} & \multicolumn{4}{c}{\textbf{RAG (best retriever)}} \\
\cmidrule(lr){2-5}\cmidrule(lr){6-9}
 & \multicolumn{2}{c}{\textsc{RebuttalSourced}} & \multicolumn{2}{c}{\textsc{LLMJudged}}
 & \multicolumn{2}{c}{\textsc{RebuttalSourced}} & \multicolumn{2}{c}{\textsc{LLMJudged}} \\
\cmidrule(lr){2-3}\cmidrule(lr){4-5}\cmidrule(lr){6-7}\cmidrule(lr){8-9}
\textbf{Model} & \textbf{Acc} & \textbf{F1} & \textbf{Acc} & \textbf{F1}
              & \textbf{Acc} & \textbf{F1} & \textbf{Acc} & \textbf{F1} \\
\midrule
\texttt{o4-mini}          & 0.245 & 0.246 & 0.442 & 0.416 & 0.244$_\text{D}$  & 0.236 & 0.514$_\text{D}$  & 0.454 \\
\texttt{GPT-5-mini}       & 0.280 & 0.275 & 0.379 & 0.381 & 0.294$_\text{D}$  & 0.279 & 0.414$_\text{D}$  & 0.399 \\
\texttt{Claude-Haiku}     & \textbf{0.295} & 0.294 & 0.295 & 0.293 & 0.297$_\text{DR}$ & 0.275 & 0.374$_\text{DR}$ & 0.336 \\
\texttt{Qwen3-8B}         & 0.276 & 0.268 & 0.171 & 0.192 & 0.287$_\text{D}$  & 0.271 & 0.371$_\text{D}$  & 0.336 \\
\texttt{Qwen2.5-7B}         & 0.278 & 0.269 & 0.170 & 0.190 & \textbf{0.324}$_\text{D}$ & 0.277 & 0.209$_\text{D}$ & 0.200 \\
\bottomrule
\end{tabular}}

\label{tab:macro-f1}
\end{table*}

Table~\ref{tab:macro-f1} reports accuracy and macro-F1 for both full-context and RAG paradigms. In the RAG setting, \texttt{Qwen2.5-7B} Dense achieves the highest \textsc{RebuttalSourced} accuracy (0.324, macro-F1 0.277), followed by \texttt{GPT-5-mini} Dense (Acc 0.294, F1 0.279) and \texttt{o4-mini} Dense (Acc 0.244, F1 0.236). The large accuracy--macro-F1 gaps for some models reflect calibration bias rather than genuine reasoning, motivating macro-F1 as a complementary metric robust to label imbalance.

\subsection{PaperSourced RAG Results}
\label{app:b1-rag}

Table~\ref{tab:rq2-b1} reports the \textsc{PaperSourced} Oracle upper bound alongside Dense+R for all five models; the full four-retriever \textsc{PaperSourced} results are in the main RAG table (Table~\ref{tab:rag-b2-b6}). Oracle feeds the gold-standard supporting paragraph directly to the verifier, providing an upper bound under perfect retrieval.

The Oracle upper bound reveals that \textsc{PaperSourced} is nearly solvable when retrieval is perfect: \texttt{o4-mini} reaches 0.988, \texttt{GPT-5-mini} 0.958, \texttt{Qwen3-8B} 0.872, and \texttt{Claude-Haiku} 0.860. The gap between Oracle and Dense+R directly quantifies accuracy lost to retrieval imperfection: 0.084 for \texttt{o4-mini} (0.988$\to$0.904) and 0.106 for \texttt{GPT-5-mini}. One anomaly: \texttt{Qwen2.5-7B} Dense+R (0.860) exceeds its own Oracle (0.816), because Dense+R supplies ten passages (including adjacent context) while Oracle supplies only the single gold paragraph. For a model with limited long-context capacity, richer surrounding context improves its ability to confirm supported claims even when the exact evidence is present in both conditions.

\subsection{Extended RAG Analysis}
\label{app:rag-extended}

This section collects secondary RAG findings deferred from Section~\ref{sec:rag_Res}.

\paragraph{Performance gap narrows on interpretive claims.}
On \textsc{RebuttalSourced}, all retrieval configurations cluster between 0.234 and 0.244 for \texttt{o4-mini}, closely tracking the full-context result of 0.245. For \textsc{LLMJudged}, the picture inverts: \texttt{Qwen2.5-7B} accuracy collapses to 0.209--0.228 with macro-F1 close to accuracy (e.g., 0.200 for Dense), confirming that its \textsc{RebuttalSourced} accuracy advantage does not transfer to semantically harder claims.

\paragraph{\texttt{Qwen2.5-7B} collapses on \textsc{LLMJudged} yet leads on \textsc{Human-Verified}.}
On \textsc{LLMJudged}, among the models scored by re-scoring \textsc{RebuttalSourced} predictions, \texttt{GPT-5-mini} leads (best: BM25-R 0.417), while \texttt{Qwen2.5-7B} collapses to 0.209--0.228, a 2$\times$ gap confirming a hard capability boundary on semantic entailment. On \textsc{Human-Verified}, \texttt{Qwen2.5-7B} recovers sharply: BM25-R reaches 0.367, the highest \textsc{Human-Verified} score across all models and retrievers, consistent with its tendency to over-predict Supported and Partially Supported, the dominant categories in the human-annotated set.

\paragraph{Retriever choice matters most for grounded claims.}
Across \textsc{PaperSourced} and \textsc{verifiable-only}, BM25+Reranker is usually the strongest retriever, beating pure BM25 and dense retrieval for every model on \textsc{PaperSourced} and for most on \textsc{verifiable-only} (the exception is \texttt{GPT-5-mini}, where dense retrieval is marginally higher). Reranking improves precision at low ranks, which is critical when the verifier's context is restricted to a small passage set. For \textsc{LLMJudged}, dense retrieval surpasses the reranker for \texttt{o4-mini} (0.514 vs.\ 0.499), suggesting that semantic embeddings better capture paraphrased or implicit evidence that lexical matching misses.

\subsection{Classical Non-LLM Baselines}
\label{app:nli}

A fair question about any LLM-based system is whether the LLM is doing work that a smaller, cheaper model could do. We tested this directly with three zero-shot entailment models trained on MultiNLI \citep{williams2018mnli}: RoBERTa-large-MNLI \citep{liu2019roberta}, BART-large-MNLI \citep{lewis2020bart}, and DeBERTa-large-MNLI \citep{he2021deberta}. For each claim, the premise is the concatenation of the top-3 retrieved chunks and the hypothesis is the claim, and the three-way output is mapped onto our schema (entailment to \texttt{Supported}, contradiction to \texttt{Not Supported}, neutral to \texttt{Not Determined}). We ran every model under all four retrieval configurations, so that a weak result could not be attributed to one bad retriever.

\begin{table*}[t]
\centering
\caption{Zero-shot NLI baselines under all four retrieval configurations, reported as Accuracy / macro-F1 on all six \system benchmarks. \textbf{Best LLM} is the single highest-accuracy model--retriever pair per benchmark, taken from Table~\ref{tab:rag-b2-b6}. No NLI configuration exceeds a macro-F1 of $0.24$ on any benchmark, and the ranking is stable across retrievers, so the gap is a property of the task formulation rather than of a particular retriever. $^{\ddagger}$NLI baselines and the matched \textbf{Best LLM} reference on \textsc{Human-Verified} are computed on the original 150-instance subset.}
\resizebox{\textwidth}{!}{%
\begin{tabular}{llcccccc}
\toprule
\textbf{Method} & \textbf{Retriever}
  & \makecell{\textsc{Paper}\\\textsc{Sourced}}
  & \makecell{\textsc{Rebuttal}\\\textsc{Sourced}}
  & \makecell{\textsc{LLM}\\\textsc{Judged}}
  & \makecell{\textsc{High}\\\textsc{Agreement}}
  & \makecell{\textsc{Verifiable}\\\textsc{-only}}
  & \makecell{\textsc{Human}\\\textsc{Verified}$^{\ddagger}$} \\
\midrule
\multirow{4}{*}{RoBERTa-MNLI}
  & BM25            & 0.070 / 0.044 & 0.147 / 0.116 & 0.401 / 0.212 & 0.234 / 0.162 & 0.206 / 0.152 & 0.167 / 0.106 \\
  & BM25+Reranker   & 0.072 / 0.045 & 0.136 / 0.108 & 0.386 / 0.205 & 0.206 / 0.143 & 0.168 / 0.126 & 0.147 / 0.103 \\
  & Dense           & 0.066 / 0.041 & 0.145 / 0.121 & 0.384 / 0.206 & 0.206 / 0.152 & 0.168 / 0.132 & 0.153 / 0.120 \\
  & Dense+Reranker  & 0.068 / 0.042 & 0.136 / 0.108 & 0.390 / 0.205 & 0.194 / 0.137 & 0.168 / 0.131 & 0.147 / 0.106 \\
\midrule
\multirow{4}{*}{BART-MNLI}
  & BM25            & 0.426 / 0.199 & 0.176 / 0.129 & 0.359 / 0.232 & 0.303 / 0.224 & 0.260 / 0.204 & 0.153 / 0.124 \\
  & BM25+Reranker   & 0.490 / 0.219 & 0.191 / 0.143 & 0.356 / 0.230 & \textbf{0.326} / \textbf{0.238} & 0.275 / 0.213 & 0.167 / 0.134 \\
  & Dense           & 0.372 / 0.181 & 0.175 / 0.134 & 0.367 / \textbf{0.238} & 0.314 / 0.236 & 0.275 / 0.218 & 0.173 / \textbf{0.145} \\
  & Dense+Reranker  & \textbf{0.512} / \textbf{0.226} & 0.188 / 0.142 & 0.351 / 0.228 & 0.314 / 0.237 & 0.267 / 0.215 & 0.173 / 0.141 \\
\midrule
\multirow{4}{*}{DeBERTa-MNLI}
  & BM25            & 0.184 / 0.104 & 0.166 / 0.133 & 0.406 / 0.226 & 0.280 / 0.205 & 0.229 / 0.178 & \textbf{0.187} / 0.140 \\
  & BM25+Reranker   & 0.216 / 0.118 & 0.149 / 0.121 & \textbf{0.422} / 0.233 & 0.257 / 0.186 & 0.214 / 0.166 & 0.147 / 0.114 \\
  & Dense           & 0.176 / 0.100 & 0.136 / 0.109 & 0.374 / 0.198 & 0.211 / 0.153 & 0.206 / 0.162 & 0.133 / 0.105 \\
  & Dense+Reranker  & 0.290 / 0.150 & 0.135 / 0.105 & 0.400 / 0.221 & 0.240 / 0.176 & 0.191 / 0.146 & 0.153 / 0.118 \\
\midrule
\textbf{Best LLM} & best per benchmark
  & 0.904 / 0.237 & 0.324 / 0.157 & 0.514 / 0.454 & 0.514 / 0.499 & 0.519 / 0.486 & 0.367 / 0.300 \\
\bottomrule
\end{tabular}%
}
\label{tab:nli-baselines}
\end{table*}

Table~\ref{tab:nli-baselines} reports all 72 configurations. Macro-F1 never exceeds 0.24 on any benchmark under any retriever, against 0.45 to 0.50 for the best LLM configuration on the document-decidable splits. Two structural causes explain the gap, and neither is a tuning problem. Three-way NLI has no output for \texttt{Partially Supported}, which is the single largest class in \textsc{RebuttalSourced} at 310 of 800 claims, so the ceiling is imposed by the label space before inference begins. And MNLI models cap the premise at 512 tokens, which truncates most of a 10-chunk evidence set, so the model frequently decides on evidence it never read.

The accuracy column occasionally looks less bad than the macro-F1 column, and that discrepancy is itself informative. BART-MNLI reaches 0.512 accuracy on \textsc{PaperSourced} with a macro-F1 of 0.226, and RoBERTa-MNLI reaches 0.401 accuracy on \textsc{LLMJudged} at 0.212 macro-F1. Both come from predicting one label heavily on a skewed split, the same distribution-matching effect we discuss for \texttt{Qwen2.5-7B} in Appendix~\ref{app:ps-preddist}. Read on macro-F1, the ranking is stable: off-the-shelf entailment is not a viable substitute here, and the value of the LLM verifiers lies in long-context, four-way reasoning performed zero-shot, not in raw entailment ability.

\subsection{Error Analysis on \textsc{PaperSourced}}
\label{app:error-analysis}

\textsc{PaperSourced} is the only split where we know the correct label for every instance without appeal to a judge, since each claim is extracted from a manuscript passage and is genuinely \texttt{Supported} by construction. That makes it a clean setting for asking what the remaining errors actually are, rather than only how many there are. We sampled 150 errors, stratified as 30 per verifier under the RAG setting, and read each one against the retrieved evidence and the source passage. Table~\ref{tab:error-taxonomy} gives the breakdown; the categories are described below.

\begin{table}[t]
\centering
\caption{Error taxonomy on \textsc{PaperSourced} under RAG, from manual inspection of a stratified sample of 150 errors (30 per verifier). Every claim in this split is genuinely \texttt{Supported}, so any other predicted label counts as an error.}
\resizebox{\columnwidth}{!}{%
\begin{tabular}{lc}
\toprule
\textbf{Error type} & \textbf{Share} \\
\midrule
Overly strict specificity (downgrade to \texttt{Part.\ Supp.}) & 55.3\% \\
Retrieval and evidence localization miss                        & 36.0\% \\
Verifier output or parse failure                                & \phantom{0}8.7\% \\
\bottomrule
\end{tabular}%
}
\label{tab:error-taxonomy}
\end{table}

\begin{itemize}[leftmargin=1.5em, itemsep=3pt, topsep=3pt]
  \item \textbf{Overly strict specificity (55.3\%).} The verifier retrieves the right passage, recognizes the substance of the claim, and then downgrades it to \texttt{Partially Supported} because one detail is not restated verbatim: an exact count, a pointer to a theorem or appendix, a numeric threshold. The evidence supports the claim; the grader is stricter than the label definition intends.
  \item \textbf{Retrieval and localization miss (36.0\%).} The supporting passage is not among the top-10 retrieved chunks, so the verifier correctly reports that the evidence in front of it does not contain the claim and hedges to \texttt{Not Determined}. This is a retrieval failure presented as a verification error, and it is consistent with Recall@10 staying below 0.55 for every retriever (Appendix~\ref{app:retrieval}).
  \item \textbf{Output or parse failure (8.7\%).} The verifier emits a malformed response that cannot be mapped to a label, and the harness defaults to \texttt{Not Determined}.
\end{itemize}

Taken together, 91.3\% of the error mass is strictness or retrieval rather than faulty inference on evidence the model actually saw. The practical consequence is that raw \textsc{PaperSourced} accuracy overstates the reasoning error rate, and that two different fixes are indicated: better retrieval for the second category, and a calibration or rubric adjustment for the first. Neither requires a stronger reasoner.

\subsection{Predicted-Label Distribution on \textsc{PaperSourced}}
\label{app:ps-preddist}

\texttt{Qwen2.5-7B} outperforming \texttt{GPT-5-mini} on \textsc{PaperSourced} under RAG looks anomalous next to the rest of the results. It is not, and the mechanism is worth making explicit because it applies to any single-class evaluation.

\begin{table}[t]
\centering
\caption{Predicted-label distribution (\%) on \textsc{PaperSourced} under RAG, aggregated over all four retrievers. Every gold label in this split is \texttt{Supported}, so this is a single-row confusion matrix and the \texttt{Supported} column equals accuracy.}
\resizebox{\columnwidth}{!}{%
\begin{tabular}{lcccc}
\toprule
\textbf{Model} & \textbf{Supported} & \textbf{Part.\ Supp.} & \textbf{Not Det.} & \textbf{Not Supp.} \\
\midrule
\texttt{GPT-5-mini} & 83.1 & 9.5 & 5.1 & 2.3 \\
\texttt{Qwen2.5-7B} & \textbf{85.2} & 7.0 & 5.4 & 2.4 \\
\bottomrule
\end{tabular}%
}
\label{tab:ps-preddist}
\end{table}

Every gold label in \textsc{PaperSourced} is \texttt{Supported}, so the predicted-label distribution is a one-row confusion matrix and the \texttt{Supported} column \emph{is} the accuracy. Table~\ref{tab:ps-preddist} shows the two models side by side, aggregated over all four retrievers. \texttt{Qwen2.5-7B} says \texttt{Supported} 85.2\% of the time; \texttt{GPT-5-mini} says it 83.1\% of the time and hedges to \texttt{Partially Supported} or \texttt{Not Determined} more often. The 2.1-point difference is the whole margin. On a split where the correct answer is always the confident one, willingness to commit is rewarded and caution is punished, even when caution is the better-calibrated behavior. Two claims from our sample illustrate the pattern: for a claim quoting an exact mean IoU of 44.91\%, \texttt{GPT-5-mini} returns \texttt{Not Determined} because it will not certify the precise figure, and for a claim describing several cache-compression techniques evaluated under five settings it returns \texttt{Partially Supported} because it reads the enumeration as incomplete. Both are gold \texttt{Supported}.

The same prior is costly wherever the label set is not degenerate. In full-context evaluation, \texttt{GPT-5-mini} leads \texttt{Qwen2.5-7B} 0.379 to 0.170 on \textsc{LLMJudged}, 0.513 to 0.213 on \textsc{HighAgreement}, and 0.518 to 0.250 on \textsc{verifiable-only} (Table~\ref{tab:rq1-fullpaper}). So the \textsc{PaperSourced} ordering reflects a \texttt{Supported}-prior meeting an all-\texttt{Supported} split, not stronger verification, and it is the reason we report macro-F1 alongside accuracy throughout.

\subsection{Benchmark Size in Context}
\label{app:size}

At 800 claims, \system\ is small next to open-domain fact-verification corpora, and it is reasonable to ask whether conclusions drawn from it are stable. The relevant comparison is not to web-scale datasets but to claim-verification benchmarks with expert, evidence-grounded annotation, where size is bounded by annotation cost rather than by data availability.

\begin{table}[t]
\centering
\caption{Evaluation size and evidence scope of \system\ relative to expert-annotated scientific claim-verification benchmarks. Evidence scope matters as much as raw count: our labels are grounded in the full manuscript rather than an abstract or a single reference passage.}
\resizebox{\columnwidth}{!}{%
\begin{tabular}{lccl}
\toprule
\textbf{Dataset} & \textbf{\#Claims} & \textbf{Annotation} & \textbf{Evidence scope} \\
\midrule
MSVEC \citep{evans2023msvec}      & 200 & Expert & Reference paper \\
SciFact (test) \citep{wadden2020scifact} & 300 & Expert & Abstract \\
\midrule
\system\ (\textsc{Human-Verified}) & 300 & Human-audited & Full manuscript \\
\system\ (full)                    & 800 & LLM + human-audited & Full manuscript \\
\bottomrule
\end{tabular}%
}
\label{tab:benchmark-size}
\end{table}

Table~\ref{tab:benchmark-size} places our benchmark against that reference class. The full 800-claim set exceeds the SciFact test split (300) \citep{wadden2020scifact} and the MSVEC evaluation corpus (200) \citep{evans2023msvec}, and the hand-audited 300-claim subset matches them exactly. Larger corpora in this space, such as COVID-Fact \citep{saakyan2021covidfact} and HealthVer \citep{sarrouti2021healthver}, reach scale through crowdsourced or automatically derived labels over abstracts and web text; we deliberately traded that scale for manuscript-grounded reliability, since a label for a review claim is only meaningful with respect to the full paper the reviewer read. Annotating against a whole manuscript is substantially more expensive per claim than annotating against an abstract, which is the constraint the size reflects.

\subsection{Retrieval Quality Analysis}
\label{app:retrieval}

\paragraph{Retrieval configuration.} The dense retriever uses \texttt{all-MiniLM-L6-v2} sentence embeddings, and the cross-encoder reranker uses \texttt{cross-encoder/ms-marco-MiniLM-L-6-v2}. Manuscripts are segmented with Docling structure-aware (section/heading-based) chunking, producing variable-length chunks with no fixed token window (median 133 words per chunk, mean 132, p90 188). For each claim we retrieve an initial pool of $k\!=\!20$ candidate passages; when reranking is enabled, these are reranked down to the final top-$10$ passages fed to the verifier, and non-reranked configurations directly take the top-$10$.

\begin{table*}[t]
\centering
\caption{Retrieval quality on \textsc{PaperSourced}.}

\resizebox{0.6\textwidth}{!}{%
\begin{tabular}{lcccc}
\hline
\textbf{Retriever} & \textbf{Recall@5} & \textbf{Recall@10} & \textbf{nDCG@10} & \textbf{MRR@10} \\ 
\hline
BM25 & 0.398 & 0.474 & 0.570 & 0.289 \\
BM25 + Reranker & 0.434 & 0.492 & 0.608 & 0.331 \\
Dense & 0.420 & 0.528 & 0.644 & 0.299 \\
Dense + Reranker & \textbf{0.452} & \textbf{0.530} & \textbf{0.684} & \textbf{0.346} \\
\hline
\end{tabular}}

\label{tab:retrieval-unified-qwen34b}
\end{table*}

Table~\ref{tab:retrieval-unified-qwen34b} compares four retrieval strategies on \textsc{PaperSourced}, where the ground-truth supporting section is known, enabling intrinsic evaluation. Dense+Reranker achieves the best performance across all metrics (Recall@5: 0.452, Recall@10: 0.530, nDCG@10: 0.684, MRR: 0.346), while BM25 alone is weakest.

Consistent with prior work~\citep{kamalloo2024resourcesforbier}, reranking substantially boosts BM25. BM25+Reranker surpasses standalone dense retrieval at Recall@5 (0.434 vs.\ 0.420) but lags at Recall@10 (0.492 vs.\ 0.528), suggesting that lexical matching excels at placing relevant chunks at shallow ranks while dense representations improve deeper recall. Despite this, Recall@10 remains below 0.55 across all configurations: the ground-truth section is absent from the top-10 retrieved passages in nearly half of all cases. The Oracle upper bound (Table~\ref{tab:rq2-b1}) makes this bottleneck concrete: the 8.4-point gap between Oracle (0.988) and Dense+R (0.904) for \texttt{o4-mini} is attributable entirely to retrieval imperfection.

\subsection{Token Efficiency and Operating Cost}
\label{app:token-efficiency}

RAG-based verification substantially reduces token consumption relative to full-context processing. Tables~\ref{tab:rag_tokens}, \ref{tab:fullcontext_tokens}, and~\ref{tab:comparison} report average token usage per claim across all six benchmarks in both settings.

\begin{table}[t]
\centering
\caption{Tokens per claim, RAG verification (four retrievers, six benchmarks).}
\label{tab:rag_tokens}
\resizebox{\columnwidth}{!}{%
\begin{tabular}{lcccc}
\toprule
\textbf{Benchmark} & \textbf{BM25} & \textbf{FAISS} & \textbf{Dense+R} & \textbf{BM25+R} \\
\midrule
\textsc{PaperSourced}    & 2,235.85 & 2,231.82 & 2,384.76 & 2,291.40 \\
\textsc{RebuttalSourced} & 2,156.78 & 2,128.87 & 2,371.78 & 2,236.87 \\
\textsc{LLMJudged}       & 2,156.78 & 2,128.87 & 2,371.78 & 2,236.87 \\
\textsc{HighAgreement}   & 2,112.56 & 2,085.18 & 2,262.94 & 2,205.61 \\
\textsc{verifiable-only} & 2,153.54 & 2,240.62 & 2,253.42 & 2,244.24 \\
\textsc{Human-Verified}  & 1,922.49 & 1,986.74 & 1,976.66 & 1,945.26 \\
\midrule
\textbf{Average}         & \textbf{2,123.00} & \textbf{2,133.68} & \textbf{2,270.22} & \textbf{2,193.38} \\
\bottomrule
\end{tabular}%
}
\end{table}

\begin{table}[t]
\centering
\caption{Tokens per claim, full-context verification (six benchmarks).}
\label{tab:fullcontext_tokens}
\resizebox{0.8\columnwidth}{!}{%
\begin{tabular}{lc}
\toprule
\textbf{Benchmark} & \textbf{Tokens per Claim} \\
\midrule
\textsc{PaperSourced}    & 21,648.85 \\
\textsc{RebuttalSourced} & 21,130.27 \\
\textsc{LLMJudged}       & 21,130.27 \\
\textsc{HighAgreement}   & 22,323.60 \\
\textsc{verifiable-only} & 21,333.30 \\
\textsc{Human-Verified}  & 18,912.43 \\
\midrule
\textbf{Average}         & \textbf{21,079.79} \\
\bottomrule
\end{tabular}}
\end{table}

\begin{table}[t]
\centering
\caption{Token efficiency, RAG vs.\ full-context. Reduction~=~full-context/RAG token ratio.}
\label{tab:comparison}
\resizebox{\columnwidth}{!}{%
\begin{tabular}{lccc}
\toprule
\textbf{Benchmark} & \textbf{Full-Context} & \textbf{RAG (Avg)} & \textbf{Reduction} \\
\midrule
\textsc{PaperSourced}    & 21,648.85 & 2,285.96 & 9.47$\times$ \\
\textsc{RebuttalSourced} & 21,130.27 & 2,223.57 & 9.50$\times$ \\
\textsc{LLMJudged}       & 21,130.27 & 2,223.57 & 9.50$\times$ \\
\textsc{HighAgreement}   & 22,323.60 & 2,166.57 & 10.30$\times$ \\
\textsc{verifiable-only} & 21,333.30 & 2,222.96 & 9.60$\times$ \\
\textsc{Human-Verified}  & 18,912.43 & 1,957.79 & 9.66$\times$ \\
\midrule
\textbf{Overall Average} & \textbf{21,079.79} & \textbf{2,180.07} & \textbf{9.67$\times$} \\
\bottomrule
\end{tabular}%
}
\end{table}

On average, RAG reduces token consumption by roughly $9.7\times$ relative to full-context verification, consistently across all six benchmarks, from $9.47\times$ on \textsc{PaperSourced} to $10.30\times$ on \textsc{HighAgreement}. This is what makes the pipeline runnable at venue scale rather than only on a sample, and it is also what lets open-weight models with restricted context windows participate at all. The saving does not cost accuracy on grounded benchmarks, and on \textsc{PaperSourced} it improves it: Dense+R reaches 0.904 for \texttt{o4-mini} against 0.866 in full-context, because a focused evidence set is easier to reason over than a whole manuscript.

Together with the extraction costs in Table~\ref{tab:extract-cost}, this fixes the operating cost of a deployment. Extraction runs locally at no inference cost and a median of 8.2 seconds per review, and verification consumes about 2.2K tokens per claim rather than 21K. These are the measurements behind the deployment discussion in Appendix~\ref{app:deployment}.

\subsection{Where Intrinsic Retrieval Metrics Apply}
\label{app:retrieval-scope}

Recall@$k$, nDCG@$k$, and MRR are reported only on \textsc{PaperSourced}, and the Oracle upper bound is computed there as well. This is a constraint of the data rather than a choice about what to report. All three metrics need a gold evidence span, and spans exist only where a claim was drawn from a known passage. For \textsc{RebuttalSourced}, \textsc{LLMJudged}, and \textsc{Human-Verified}, the label comes from an author--reviewer exchange or from a whole-manuscript reading, and neither procedure identifies the paragraph that settles the claim. Some claims are settled by an absence, such as ``no ablation is reported,'' which has no supporting span anywhere in the document. Annotating spans for those splits would mean inventing a ground truth we do not have.

The question behind the metric, whether retrieval or reasoning limits performance, can still be answered on those splits, by removing retrieval instead of measuring it. Full-context verification passes the entire manuscript to the verifier and skips the retrieval stage, so it is a retrieval-free upper bound. If retrieval were binding, accuracy should rise sharply. It does not: on \textsc{Human-Verified}, \texttt{GPT-5-mini} reaches 0.317 in full-context against 0.297 with its best retriever, and \texttt{o4-mini} reaches 0.267 against 0.257 (Tables~\ref{tab:rq1-fullpaper} and~\ref{tab:hv-rag-300}). Perfect retrieval buys a point or two, so what limits these splits is verification reasoning. \textsc{PaperSourced} behaves in exactly the opposite way, where the 8.4-point Oracle-to-Dense+R gap for \texttt{o4-mini} is retrieval and nothing else, which is why we keep both regimes in the evaluation.

\section{Deployment Setting and Intended Use}
\label{app:deployment}

\paragraph{The pipeline runs at the scale of a real venue.} All three stages are implemented end to end and are cheap enough to apply to an entire submission cycle. Claim extraction runs locally on a single consumer GPU at a median of 8.2 seconds per review with no API cost (Appendix~\ref{app:extract-cost}), and retrieval-augmented verification consumes roughly $9.7\times$ fewer tokens per claim than passing the verifier a whole manuscript (Appendix~\ref{app:token-efficiency}). Because both figures are measured rather than estimated, a venue can compute the cost of a deployment in advance. This is what we mean when we describe \system\ as production-ready: the engineering is complete, the throughput is known, and nothing in the design assumes a research-scale workload.

\paragraph{Every prediction ships with its evidence.} The output of the pipeline is not a bare label. Each claim is returned together with the manuscript passages the verifier conditioned on, so a decision can always be traced back to the text that produced it. This is what makes the system useful in an editorial workflow: a reader can confirm or overturn any prediction in seconds by looking at the passages already retrieved for them, without opening the paper and searching from scratch.

\paragraph{The intended workflow is human-in-the-loop by design.} \system\ is built to be an assistant to editorial judgement rather than a replacement for it. An area chair opens a review and sees which reviewer claims the system flagged as needing a closer look, each attached to the relevant manuscript passages, and directs attention accordingly. Authors can use the same view when preparing a rebuttal, and reviewers can use it to self-check a draft review before submitting. The value the system delivers is triage and evidence retrieval at scale, which is precisely the part of the task that does not scale for humans, while the final judgement stays where it belongs.

\paragraph{The benchmark is deliberately demanding.} \system\ ships with an evaluation suite built to remain informative as models improve. The variants span the full difficulty range, from \textsc{PaperSourced}, where evidence is explicit and the best verifier reaches 0.904 under RAG, through to \textsc{RebuttalSourced}, where claims are drawn from live author--reviewer disputes and mix factual observation with interpretation. Setting the harder end of that range beyond what current frontier models saturate is a design goal rather than a shortcoming: it gives the community headroom to measure progress on manuscript-grounded verification for several model generations, and it keeps the assistive framing above honest about which claims a human should look at first.

\end{document}